\documentclass[runningheads]{llncs}

\usepackage{eccv}

\usepackage{eccvabbrv}

\usepackage{siunitx}
\usepackage{caption}
\usepackage{multirow}
\usepackage{graphicx}
\usepackage{booktabs}

\usepackage[accsupp]{axessibility}  

\usepackage{hyperref}

\usepackage{orcidlink}

\usepackage[marginal]{footmisc}

\usepackage[linesnumbered,lined,ruled,commentsnumbered]{algorithm2e}
\usepackage{algpseudocode}

\usepackage{xr}
\usepackage{tabularx}

\newcolumntype{Y}{>{\centering\arraybackslash}X}
\newcolumntype{C}[1]{>{\centering\arraybackslash}p{#1}}
\newcolumntype{L}[1]{>{\arraybackslash}p{#1}}

\begin{document}

\title{Spline-based Transformers} 


\author{
    Prashanth Chandran\inst{1}\textsuperscript{*}
    \and
    Agon Serifi\inst{2,3}\textsuperscript{*}
    \and \newline
    Markus Gross\inst{1,3}
    \and
    Moritz B{\"a}cher\inst{2}
}

\authorrunning{P.~Chandran and A.~Serifi et al.}

\institute{
    DisneyResearch|Studios, Switzerland 
    \and
    Disney Research, Switzerland \\ 
    \email{\{prashanth.chandran, moritz.baecher\}@disneyresearch.com} 
    \and
    ETH Zurich, Switzerland \\
    \email{\{agon.serifi, grossm\}@inf.ethz.ch}
}

\maketitle

\footnotetext{\textsuperscript{*} equal contribution.}

\begin{abstract}
We introduce Spline-based Transformers, a novel class of Transformer models that eliminate the need for positional encoding. 
Inspired by workflows using splines in computer animation, our Spline-based Transformers embed an input sequence of elements as a smooth trajectory in latent space. Overcoming drawbacks of positional encoding such as sequence length extrapolation, Spline-based Transformers also provide a novel way for users to interact with transformer latent spaces by directly manipulating the latent control points to create new latent trajectories and sequences. We demonstrate the superior performance of our approach in comparison to conventional positional encoding on a variety of datasets, ranging from synthetic 2D to large-scale real-world datasets of images, 3D shapes, and animations.

\end{abstract}

\section{Introduction}
\label{sec:intro}

Positional encoding is an essential component in transformer models, introduced in the seminal work by Vaswani \etal \cite{vaswani2017attention}. It infuses positional information into input tokens to help transformers learn position-agnostic token embeddings. Positional encoding works by (1) pre-assigning sinusoids of different frequencies and phases to every position an input token can take on in a sequence, and (2) by adding this sinusoid to the token embedding that appears at the corresponding position in the sequence. Injecting a token with positional information, also referred to as \textit{absolute} position encoding in later work, has evolved into several variants that address shortcomings and improve generalization. For example, several works have shown that absolute position encoding limits the ability of transformers to handle longer sequences at inference time and proposed \textit{relative} position encoding schemes where a fixed or learned bias is added to the attention matrix~\cite{raffel2020t5,press2022alibi,ke2021unitedpe}. Irrespective of their exact arrangement, today's state-of-the-art transformer architectures employ a combination of absolute and relative position encoding schemes \cite{press2022alibi, su2023roformer, ruoss2023randomized}. 

Positional encoding assumes that token embeddings represent elemental data in a collection, \eg, individual words in a sentence, images in a video, or poses in an animation, and that an additional notion of position is required to model a collection of such elements, such as sequences of words, images, or animation frames. This thought process conceptually decouples elemental and collective datatypes and forces that separate representations for the elements and the collection as a whole are learned. This becomes even more evident when we consider that most existing architectures that learn compact neural representations for collections do not leverage the fact that individual elements, traversed in a particular order, make up a collection. 

In this work, we argue that learned neural representations for elemental and collective datatypes do not have to be decoupled from one another. Instead, they can be effectively represented in a single, shared latent space. At the heart of our approach is the idea that a collection can be represented by learning to traverse a trajectory in the latent space of elemental data. Inspired by animation workflows where splines are commonly used to describe a temporal sequence of poses, we introduce a new class of transformer models based on splines that we call \emph{Spline-based Transformers}. They do not require absolute position encoding. 

\begin{figure}
    \centering
    \includegraphics[width=\textwidth]{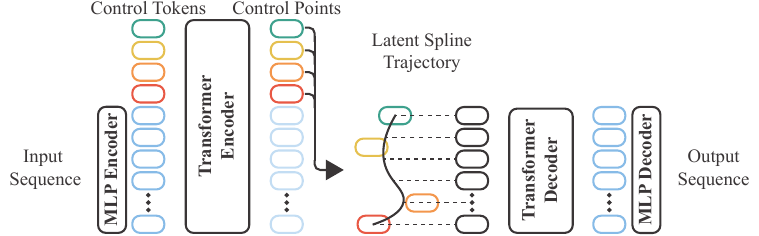}
    \caption{\textbf{Spline-based Transformers.} Our Spline-based Transformers encode an input sequence, together with learnable control tokens, into a trajectory in latent space defined by the latent control points of a spline curve.}
    \label{fig:overview}
\end{figure}

At a high level, our approach uses a transformer-based encoder with additional learned control tokens to reduce an input collection of elements to a fixed number of latent control points $\in \mathbb{R}^d$. These control points are interpreted as the control points of a $d$-dimensional spline in latent space, representing a continuous latent space trajectory. The trajectory encapsulates the fundamental characteristics of the elements constituting the input collection. Uniformly sampling and processing the trajectory through the transformer-based decoder reconstructs the original input sequence. Our Spline-based Transformers require no sinusoidal positional encoding and, therefore, completely circumvent the downsides of absolute position encoding, including poor extrapolation and overfitting. A conceptual overview of our approach is illustrated in \cref{fig:overview}.

We demonstrate the superior performance of our Spline-based Transformers over transformers with conventional positional encoding on several datasets and applications, including synthetic data (\cref{subsec:synthetic}), images (\cref{subsec:images}), animation data (\cref{subsec:animation}), and in representing challenging geometry like hair strands (\cref{subsec:strands}). Additionally, our Spline-based Transformers allow users to manipulate a given collection by directly interacting with corresponding latent controls, thereby introducing a new means of interacting with this architecture. With transformers gaining significant attention in recent years as general purpose architectures \cite{dosovitskiy2021image, radford2023speech, petrovich21actor, chandran2022shape, Aneja2023facetalk, Peebles2022DiT}, we believe that our simple yet effective approach has the potential to be leveraged across multiple disciplines for a wide variety of tasks.

Succinctly, our contributions are:
\begin{itemize}
    \item We introduce Spline-based Transformers; Transformer models that use a spline-based latent space to encode temporal information without requiring additional positional encoding.
    \item We show that simple control mechanisms to manipulate the latent space are automatically learned by our models and allow for rapid manipulation of the output sequence.
    \item We demonstrate superior performance of Spline-based Transformers over Transformers with positional encoding on a variety of data modalities, including synthetic data, images, 3D shapes, and motion datasets.
\end{itemize}
\section{Related Work}

Before we discuss related work, we would like to clarify our use of the term positional encoding. The term positional encoding is widely used in the literature of coordinate-based neural networks \cite{mildenhall2022nerf, Park_2019_CVPR} to refer to a frequency encoding scheme for improvement of network training~\cite{tancik2020fourfeat} wherein a low dimensional input (such as a 3D position) is mapped to a higher dimension using a collection of sinusoids of different frequencies. In our work, we refer to positional encoding as a mechanism to \emph{introduce} positional information into inputs and outputs that are otherwise devoid of any positional information, as is common in the literature on transformers.

Transformers~\cite{vaswani2017attention} were introduced as an alternative to traditional sequence models such as RNNs~\cite{rumelhart1986rnn} and LSTMs~\cite{hochreiter1997lstm}. While they initially showed remarkable performance on language tasks, their effectiveness as a general purpose neural architecture led to their quick adoption as foundational image models~\cite{dosovitskiy2021image, liu2021Swin, liu2021swinv2}, in speech recognition~\cite{radford2023speech}, 3D and 4D modeling~\cite{chandran2022shape, Aneja2023facetalk, petrovich21actor, chandran2022facial}, and more recently as backbone architectures for diffusion models~\cite{Peebles2022DiT}. 

The original transformers~\cite{vaswani2017attention} relied on absolute position encoding with sinusoids to inject positional and temporal information into input tokens. Soon after, researchers identified sequence length extrapolation and overfitting as limitations of absolute positional encoding and introduced several extensions to combat them. Su \etal proposed \emph{RoPE}~\cite{su2023roformer}, where absolute positions are encoded as a rotation matrix and relative token positions are explicitly taken into account during attention computations for better performance and generalization. Raffel \etal introduced the T5 transformer~\cite{raffel2020t5} where a learned bias that depends on the relative distance between tokens is added to the attention matrix. They achieved a performance boost on a variety of natural language tasks. Press \etal proposed \emph{ALiBi}~\cite{press2022alibi} and showed that a fixed bias with a predetermined slope that depends on the attention head can improve the performance for unseen sequence lengths. An extension to ALiBi~\cite{khateeb2023pos}, which applied ideas from RoPE, further improved the performance of ALiBi in language tasks. To specifically tackle the extrapolation problem, Ruoss \etal proposed positional encoding with a randomized ordering of sinusoids~\cite{ruoss2023randomized} to account for longer test positions by augmenting the training distribution. A more recent and highly relevant finding in the context of our work is described by Kazemnejad \etal~\cite{kazemnejad2023impact}. They showed that transformers with no positional encoding (NoPE) outperform most commonly used forms of position encoding in \emph{decoder} only tasks. 

However, for transformers used in learning condensed latent spaces of sequential data~\cite{CLIP2021,tevet2022motionclip,duan2021single,petrovich21actor,chandran2022facial,chandran2022shape}, almost all current methods make use of a transformer autoencoder with relative position encoding~\cite{press2022alibi, raffel2020t5} and an additional [CLS] (short for classification) token~\cite{Bert2018} as input. In these applications, the transformer encoder aggregates information from the input data tokens into the [CLS] token, which is then interpreted as a condensed latent descriptor of the input sequence at the encoder's output. This latent descriptor token is appropriately position-encoded and passed through a transformer decoder to reconstruct the input. For such scenarios, the use of no positional encoding (NoPE) is not a viable solution as it reduces to an $n$-fold duplication of the latent descriptor, therefore passing an identical or static latent sequence to the decoder. Without any variation in its input or absolute positional encoding, the decoder fails to reconstruct the original input sequence. 

Our Spline-based Transformers present a new approach to learning such condensed latent spaces for sequential data using a transformer autoencoder that does not require a positional encoding. In addition to providing significant performance benefits, our approach provides a novel control mechanism to navigate the latent spaces without any additional complexity.  
\section{Spline-based Transformers}

The core of our architecture is constructed around a transformer autoencoder model, which incorporates a latent space between the encoder and decoder components. In the following, we first introduce the theory behind the modifications that result in our Spline-based Transformers and later describe the architectural details of the new transformer autoencoder. See \cref{fig:overview} for an illustration.

\subsection{Background: Splines}

Splines have seen widespread use in function approximation, computer-aided design, and the specification and editing of animation curves in computer graphics. They provide a means to define a curve or a trajectory with a discrete set of control points and have many desirable properties. Adjustments to control points have only a local effect, and the degree of the polynomial basis provides users with control over the smoothness of a curve.

While our modeling is agnostic to the specific spline representation, we use B-Splines in our current transformer model as they provide a good trade-off between ease of implementation and fine-grained control over shape and smoothness. A B-Spline curve is a linear combination of control points, $\mathbf{p}_i$, and basis functions, $N_{i,k}(t)$, 
\begin{equation}
\mathbf{s}(t) = \sum\limits_{i=0}^n N_{i,k}(t)\mathbf{p}_i \, \, \, \, \text{for}  \, \, \, \, t \in [t_{k-1}, t_{n+1}]
\label{eq:spline},
\end{equation}
and describes a piecewise polynomial curve where each segment has degree $k$~\cite{Farin2001}. The smoothness at the interface of pairs of segments is determined by the knot vector
\begin{equation}
T = (t_0, t_1, \hdots,t_{k-1}, t_k, t_{k+1}, \hdots, t_{n-1}, t_n, t_{n+1}, \hdots, t_{n+d}).
\end{equation}

Note that a B-Spline curve does, in general, \emph{not} pass through the two end control points. Only if a knot has multiplicity $k-1$, the corresponding control point will lie on the curve, reducing the continuity at that point to $\mathcal{C}^0$. If we increase the multiplicity of a knot to $k$, the curve is $\mathcal{C}^{-1}$ and therefore discontinuous. In more general terms, a knot with multiplicity $m$ results in a curve that is $k-m-1$-differentiable, and hence $C^{k-m-1}$, at the knot. We can always normalize the time interval so that $t \in [0, 1]$.

Splines have many desirable properties, notably:
\begin{itemize}
    \item \emph{Local support:} A knot span, $t_i \leq t \leq t_{i+1}$, is only affected by $k$ control points, and a control point only has an effect on $k$ spans. Adjustments to a control point, $\mathbf{p}_i$, have an effect on the curve between $t_i$ and $t_{i+k}$. 
    \item \emph{Smoothness:} If the multiplicity of a knot is zero, a B-Spline curve is $\mathcal{C}^{k-1}$ and $k-1$-differentiable. To increase the smoothness of the curve, we can always increase the degree of the polynomial basis. 
    \item \emph{Numerical Stability:} The theory behind B-Splines is well-understood, and numerically stable and efficient algorithms exist to evaluate them~\cite{Farin2001}.
\end{itemize}

\subsection{Network Architecture}

Typically, the encoder of a transformer autoencoder reduces an input sequence of tokens into a single latent code. However, because transformers are sequence-to-sequence architectures, they require additional pooling mechanisms to condense information from the entire input sequence into a single latent token~\cite{Bert2018,attnPooling2021}. This is usually accomplished by concatenating an additional learned token to the input sequence, and by using only the latent representation of this token as input to subsequent neural networks (\eg, a decoder) or by directly using it in a training objective~\cite{CLIP2021}. The other outputs of the transformers are discarded. In classification tasks, the learned token is often referred to as the [CLS] token~\cite{Bert2018}. Finally, in order to decode the latent [CLS] token into an output sequence, it is duplicated, positional encoded appropriately, and passed through a transformer decoder that predicts the output sequence. 

Instead of only appending a single [CLS] token to the input, Spline-based Transformers append a collection of ordered control tokens to the input sequence. Specifically, Spline-based Transformers append $n+1$ control tokens to the input sequence to obtain $n+1$ control points, $\mathbf{p}_i$, that will be used to evaluate a latent spline at the output of the encoder with polynomial basis of order $k$. Latent codes corresponding to each output token are produced by evaluating the spline at the token's position according to~\cref{eq:spline}. 

\begin{figure}
    \centering
    \includegraphics[width=\textwidth]{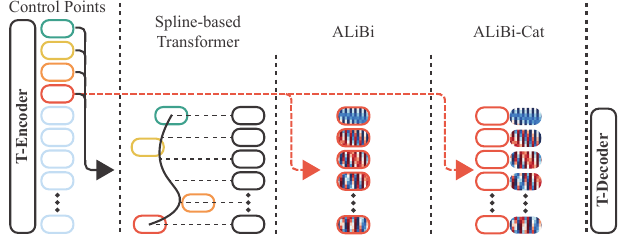}
    \caption{\textbf{Variations of Latent Spaces.} Our \textit{Spline-based Transformers} use multiple control points to evaluate a latent B-Spline and to create a $d$ dimensional trajectory in the model's latent space. On the other hand, \textit{ALiBi} duplicates a single control point and adds positional information to the duplicated points, while the positional information is concatenated to the duplicated control point in \textit{ALiBi-Cat}.}
    \label{fig:latentspace}
\end{figure}

The resulting trajectory, $\mathbf{s}(t)$, in latent space, has several advantages compared to a traditional positional encoding. First, the latent code is not perturbed by positional information, meaning the decoder does not need to learn to distinguish between positional and contextual information. Second, when using sinusoidals to encode the position of tokens, the contextual part of the token remains fixed and therefore provides a form of redundancy; our latent spline trajectories encode the temporal information implicitly, \eg, they can traverse the latent space faster in certain points and slower in others, making better use of the latent space. In \cref{fig:latentspace}, we show an overview of how our spline-based latent trajectories are derived from the control points and how they differ from commonly used schemes like ALiBi \cite{press2022alibi}.

\paragraph{Architecture Details}

As seen in \cref{fig:overview}, an input sequence is encoded using an MLP that is shared across input tokens, leading to an embedded sequence. Learnable control tokens are concatenated to the embedded sequence and sent through a seq2seq transformer encoder block. We add a linear layer after the last transformer encoder block to map the encoded tokens to the latent space dimension $d$. The exact number of evaluations of the spline depends on the number of output tokens expected at the decoder's output. Our transformer decoder uses the same structure as our encoder. The encoder and decoder have $n$-layers, each layer has $h$ heads, and $c$ feature dimensions. In every layer of the transformer, an ALiBi attention bias is added. Each layer, except the last MLP of the decoder, uses the GeLU activation~\cite{hendrycks2016gelu}. While our transformer blocks follow the structure of the T5 transformer model~\cite{raffel2020t5}, any transformer block could be used in combination with the spline-based latent space. Depending on the complexity of the data type, we use transformer blocks of varying feature dimensions and capacities. For training our Spline-based Transformer autoencoder, we use the RAdam optimizer~\cite{liu2020radam} with a cosine annealing learning rate scheduler~\cite{loshchilov2017sgdr}.

\section{Experiments}

We now present experiments on a number of datasets to demonstrate the effectiveness of the proposed Spline-based Transformers when applied to multiple modalities of sequential data. We specifically compare our method against ALiBi \cite{press2022alibi}, a state-of-the-art transformer model that uses a combination of both absolute and relative positional encoding. Because ALiBi adds sinusoids to the input token embedding, which could create an ambiguity between the token's content and position for the transformer decoder to disentangle, we also compare against a variation of ALiBi, where the sinusoids are concatenated with the token embedding, effectively doubling the size of the transformer decoder blocks. We refer to this concatenated variation as ALiBi-Cat. \cref{fig:latentspace} illustrates the differences between our spline-based latent space and the two baselines; ALiBi uses a single control point and adds positional information on top to create a latent sequence \cite{chandran2022facial, EMOTE2023}, while ALiBi-Cat concatenates the control point and the positional information instead.

For our experiments, we parameterize the latent space between the encoder and decoder blocks with cubic B\'ezier curves, with four control points per segment. B\'ezier curves are one instance of the B-Spline family, and provide sufficient smoothness for the applications we have studied so far. We uniformly sample the latent spline trajectory in the range $t \in [0, 1]$. We summarize the network parameters for each experiment in our supplemental material. 

\subsection{Synthetic Datasets}
\label{subsec:synthetic}

\begin{figure}[h]
    \centering
    \includegraphics[width=0.9\textwidth]{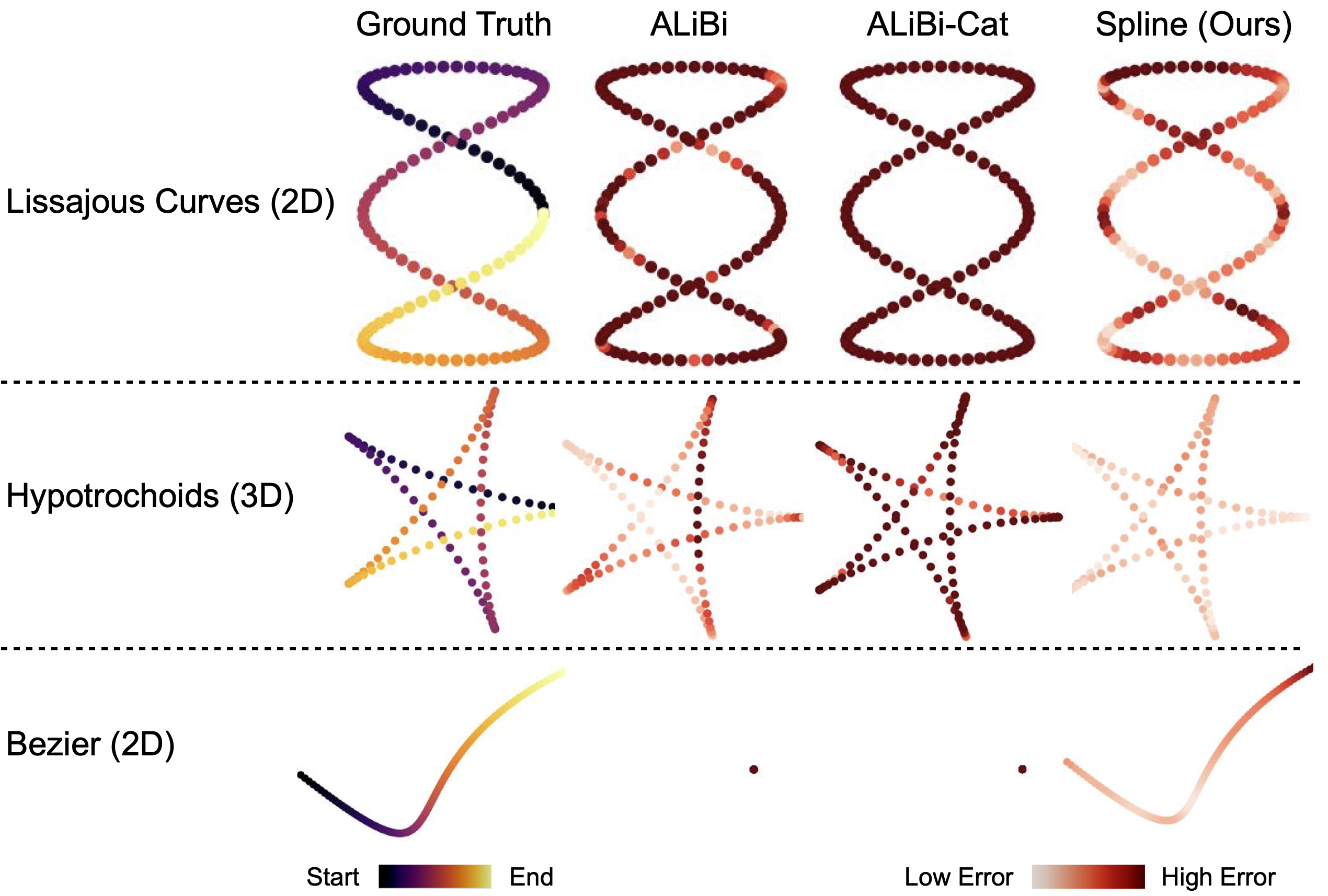}
    \caption{Our Spline-based Transformer can successfully reconstruct curves of different families with consistently better performance than ALiBi and ALiBi-Cat. In certain scenarios (third row), reconstructions from ALiBi and ALiBi-Cat can collapse to a single point, while our Spline-based Transformer successfully manages to recover the input curve.}
    \label{fig:syntheticCurves}
\end{figure}

We first evaluate our Spline-based Transformer, ALiBi, and ALiBi-Cat in representing parametric 2D curves that have a known latent space size. For this task, we use three different parametric curve families: (1) Lissajous ($d = 3$), (2) Hypotrochoids ($d=4$), (3) B\'ezier curves (with $d = 2$, and $d = 64$). For each curve type, we create three different transformer autoencoders for the Spline-based Transformer, ALiBi, and ALiBi-Cat, respectively. As seen in \cref{fig:latentspace}, the network architectures for the different autoencoders are identical, with the only difference being the mechanism used to derive latent token trajectories. The dimensionality of latent token embedding is decided based on the known latent space of the curve family. We train the three transformer autoencoders independently on each curve family. For training, we randomly sample curve parameters according to the parameterization of the curve in a pre-determined domain. Using the sampled parameters, we evaluate the curve to create a sequence of 256 2D tokens that contain the $(x,y)$ coordinates of the curve, which are then fed as input to the transformer encoder. The three transformer autoencoders are trained end-to-end using a simple L2 reconstruction objective. In \cref{tab:syn_ablation}, we show the reconstruction error of each of the transformer models when presented with 10,000 unseen curves from the family it was trained on. Our Spline-based Transformer outperforms ALiBi and ALiBi-Cat, especially on low dimensional latent spaces. Some qualitative comparisons are shown in \cref{fig:syntheticCurves}. 

\begin{table}[h]
\centering
\captionsetup{justification=centering}
\caption{Average reconstruction error (MSE) of 10000 test curves in 2D}
\label{tab:syn_ablation}
\begin{tabular}{lcccc}
\hline
Method & Lissajous (3D) & Hypotrochoids (4D) & B\'ezier (2D) & B\'ezier (64D) \\ 
\hline
ALiBi & 1e-4 & 2e-3 & 1.76e-2 & 3.88e-3 \\
ALiBi-Cat & 8e-4 & 5.3e-3 & 1.78e-2 & 3.89e-3 \\
Spline (Ours) & \textbf{3e-5} & \textbf{1.4e-3} & \textbf{2e-6} & \textbf{3.87e-3} \\
\hline
\end{tabular}
\end{table}

\subsection{Images}
\label{subsec:images}

We continue by showing the effectiveness of Spline-based Transformers in reconstructing real image datasets. For the following experiment, we divide an image into distinct non-overlapping patches to create a sequence of patches similar to (Masked) Vision Transformers~\cite{dosovitskiy2021image,he2022masked}. The sequence of 2D image patches is then the input to our transformer autoencoder. We replace the MLP-Encoder in \cref{fig:overview} with a CNN-Encoder that maps each patch of size $(PS, PS, 3)$ to a $d$-dimensional latent token $(1,d)$. An image $(H, W)$ is therefore represented with a sequence of size $(HW/PS^2,d)$. The rest of the transformer autoencoder remains identical to what was described above. We train the transformer autoencoder using a simple L2 reconstruction loss to recover the input image from the patch sequence and compare the performance of the Spline-based Transformer against ALiBi and ALiBi-Cat. 

\begin{figure}
    \centering
    \begin{subfigure}{0.48\textwidth}
        \centering
        \includegraphics[width=\textwidth]{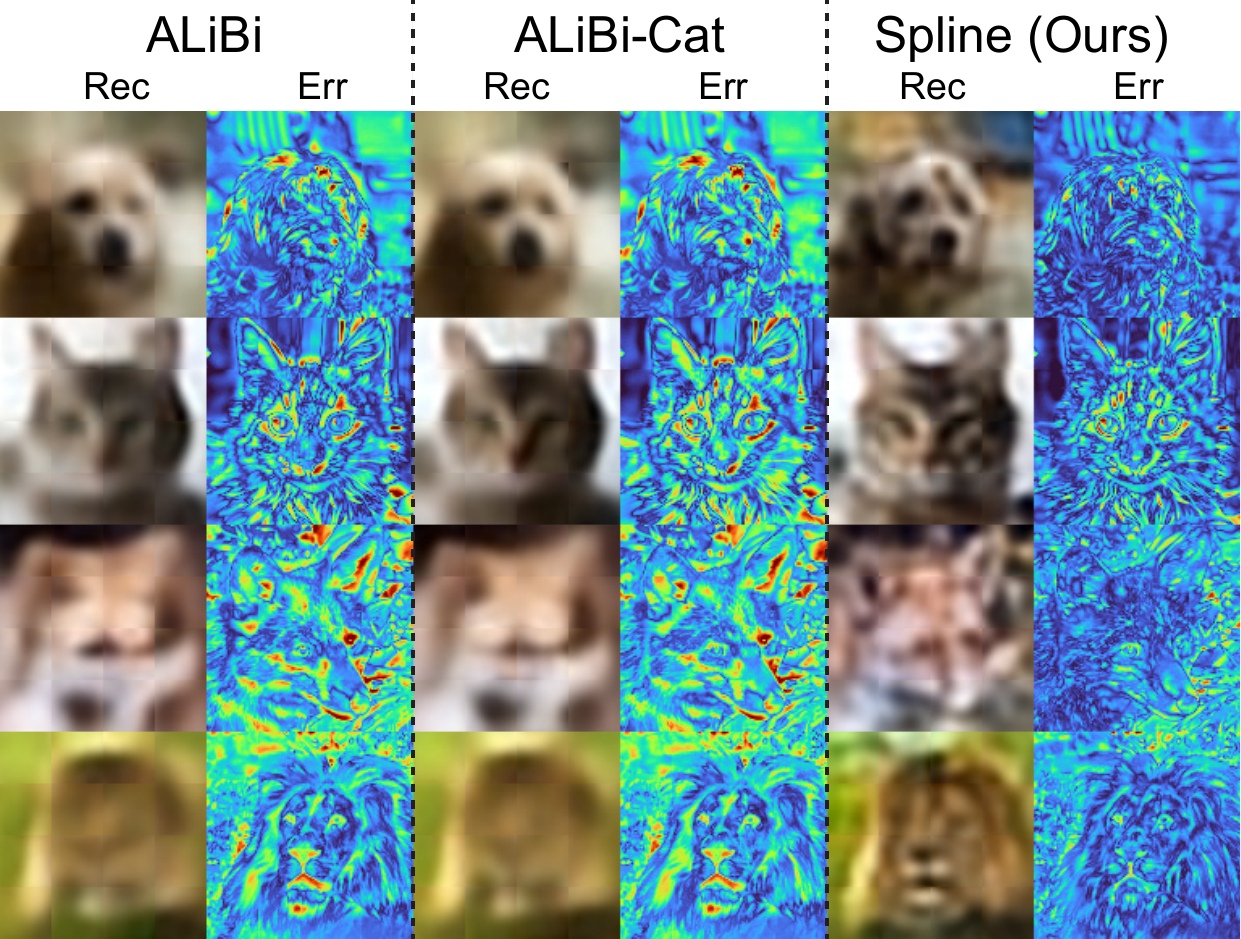}
        \caption{AFHQ - 64D}
    \end{subfigure}
    \hfill
    \begin{subfigure}{0.48\textwidth}
        \centering
        \includegraphics[width=\textwidth]{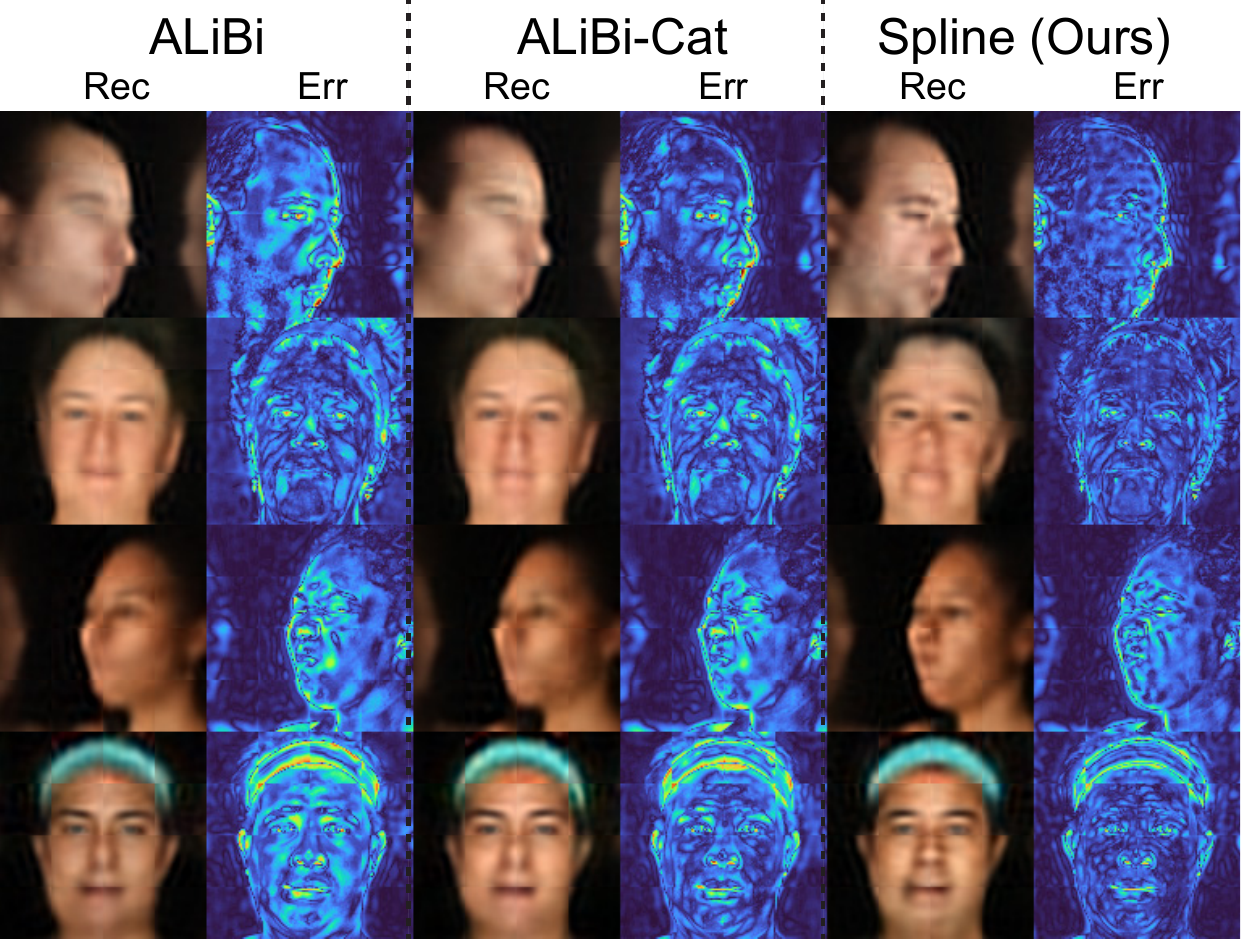}
        \caption{Faces - 64D}
    \end{subfigure}
    \caption{\textbf{Visual Image Reconstruction.} Comparison between ALiBi, ALiBi-Cat, and Spline (ours) using a 64D bottleneck on AFHQ and Faces \cite{SDFM2020}.}
    \label{fig:imageReconstruction}
\end{figure}

We present results on three different image datasets: CIFAR-10~\cite{krizhevsky2009learning} (32x32), AFHQ~\cite{choi2020starganv2} (128x128), and a dataset containing facial images~\cite{SDFM2020} (128x128). For each dataset and method, we train three transformers with three different latent sizes: 32D, 64D, and 128D. \cref{tab:img_ablation} summarizes the results. The spline-based latent space significantly outperforms the baselines by a factor of $2$. \cref{fig:imageReconstruction} shows examples of the reconstructed images and their corresponding error maps; the spline-based latent space results in sharper and more detailed images. We observe that the performance improvements are larger for lower dimensional latent spaces. More results are reported in our supplementary.

\begin{table}[h!]
\centering
\captionsetup{justification=centering}
\caption{\textbf{Image Reconstruction.} Comparison across different datasets and bottleneck dimensions. \textbf{Bold} indicates the best overall performance, and \underline{underline} the best in each category. Performance is measured in Mean Squared Error (MSE).}
\label{tab:img_ablation}
\begin{tabular}{
            @{}
            l
            >{\centering\arraybackslash}p{1cm}
            >{\centering\arraybackslash}p{1cm}
            >{\centering\arraybackslash}p{1cm}
            |>{\centering\arraybackslash}p{1cm}
            >{\centering\arraybackslash}p{1cm}
            >{\centering\arraybackslash}p{1cm}
            |>{\centering\arraybackslash}p{1.17cm}
            >{\centering\arraybackslash}p{1.14cm}
            >{\centering\arraybackslash}p{1.14cm}
            @{}
        }
\toprule
& \multicolumn{3}{c|}{CIFAR-10} & \multicolumn{3}{c|}{AFHQ} & \multicolumn{3}{c}{Faces \cite{SDFM2020}} \\
{Method} & {32D} & {64D} & {128D} & {32D} & {64D} & {128D} & {32D} & {64D} & {128D} \\
\midrule
ALiBi  & {0.266} & {0.178} & {0.107} & {0.064} & {0.050} & {0.038} & {10.65e-3} & {8.56e-3} & {6.71e-3} \\
ALiBi-Cat  & {0.264} & {0.174} & {0.108} & {0.064} & {0.049} & {0.038} & {10.87e-3} & {8.56e-3} & {7.14e-3} \\
Spline (Ours)  & {\underline{0.107}} & {\underline{0.056}} & {\textbf{0.042}} & {\underline{0.038}} & {\underline{0.030}} & {\textbf{0.025}} & {\underline{6.77e-3}}  & {\underline{5.27e-3}} & {\textbf{4.52-3}} \\
\bottomrule
\end{tabular}
\end{table}

\subsection{Animation}
\label{subsec:animation}

\begin{figure}
    \centering
    \includegraphics[width=0.9\textwidth]{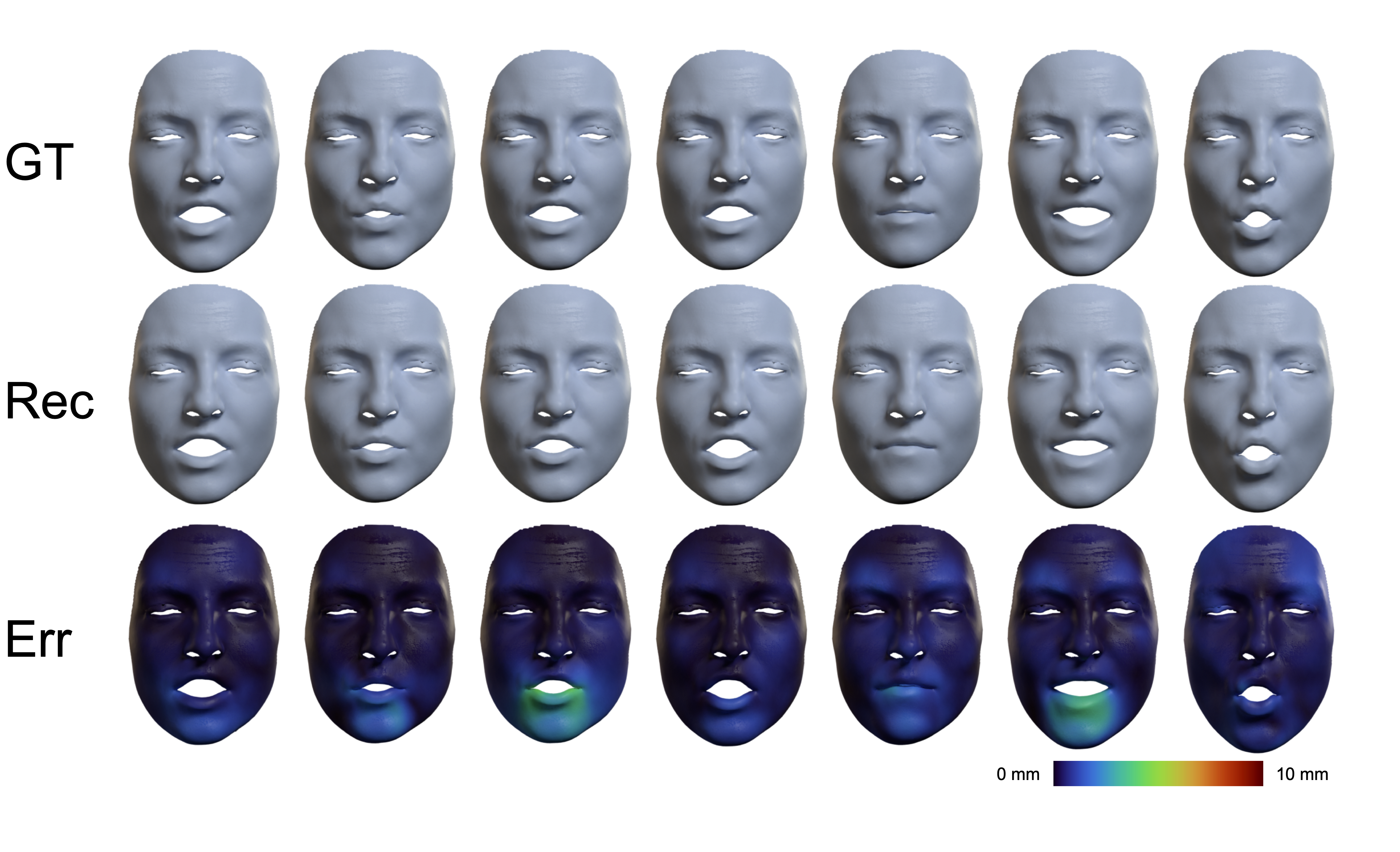}
    \caption{Our Spline-based Transformer is able to successfully represent facial animations, preserving both the identity and expression of the subject throughout the performance.}
    \label{fig:facerecon}
\end{figure}

A commonly encountered modality of sequential data is 3D animation, with interest in using transformers for learning motion manifolds garnering significant attention in recent years~\cite{petrovich21actor,chandran2022facial,Aneja2023facetalk,EMOTE2023}. A Spline-based Transformer, when used to represent 4D data, like a sequence of 3D meshes from a facial animation or a sequence of joint poses describing human motion, can lead to notable performance benefits.
\paragraph{\textbf{Faces}}
We compare the performance of the Spline-based Transformer autoencoder against ALiBi, and ALiBi-Cat, training the three models on a database of 3D facial animations~\cite{chandran2022facial}. Each animation is represented by a sequence of registered 3D meshes. We decimate them to meshes with around 5,000 vertices, and flatten the vertices to a vector. A flattened animation sequence is thereafter split into windows of size 30 ($\sim\!1$ second of animation) and used to train three variations of the transformer autoencoder. In \cref{tab:faceTable}, we show the reconstruction error on six performances from three unseen identities. Irrespective of the dimensionality of the latent space, the Spline-based Transformer outperforms both ALiBi and ALiBi-Cat. A qualitative visualization of a reconstructed test performance is shown in \cref{fig:facerecon} for the 64D Spline-based Transformer model. We note that the ALiBi transformer decoder used in these experiments is conceptually similar to the ones used in previous works~\cite{petrovich21actor,chandran2022facial,EMOTE2023,Aneja2023facetalk}, indicating that Spline-based Transformers could lead to improved performance in several downstream tasks. 

\begin{table}[h!]
\centering
\captionsetup{justification=centering}
    \caption{\textbf{Face Performance Reconstruction.} Comparison across different latent dimensions. \textbf{Bold} indicates the best overall performance, and \underline{underline} the best in each category. Performance is measured in Mean Squared Error (MSE).}
    \begin{tabular}{lcccc}
    \hline
    Method & 32D & 64D & 128D & 256D \\ 
    \hline
    ALiBi & 1.58 & 1.55 & 1.48 & 1.54 \\
    ALiBi-Cat & 1.60 & 1.54 & 1.53 & 1.50 \\
    Spline (Ours) & \underline{1.43} & \textbf{1.35} & \underline{1.47} & \underline{1.47} \\
    \hline
    \end{tabular}
    \label{tab:faceTable}
\end{table}

\paragraph{\textbf{Full-Body Motion}}

\begin{figure*}
{
\setlength{\tabcolsep}{0.8pt}
\footnotesize
\begin{tabular}{>{\centering\arraybackslash}p{0.8cm}ccccccccc}
\raisebox{23px}{GT\;} &
\includegraphics[width=0.11\textwidth]{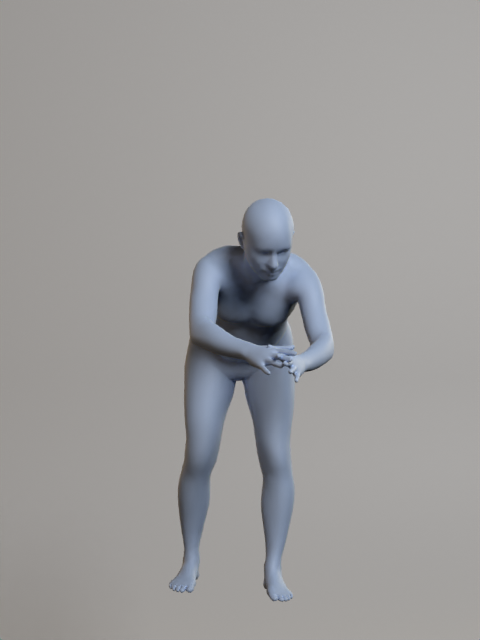} & 
\includegraphics[width=0.11\textwidth]{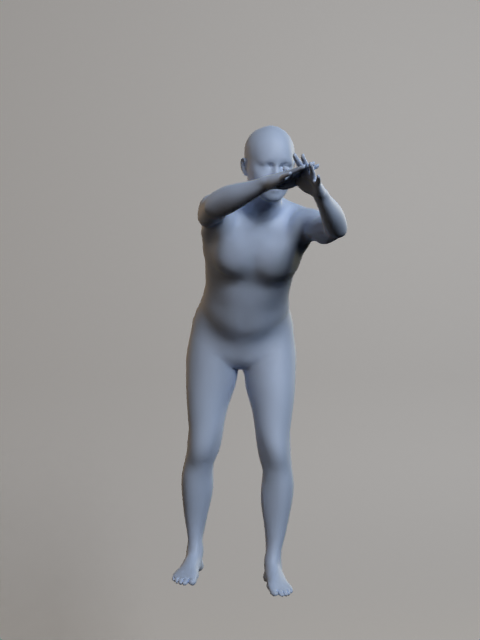} & 
\includegraphics[width=0.11\textwidth]{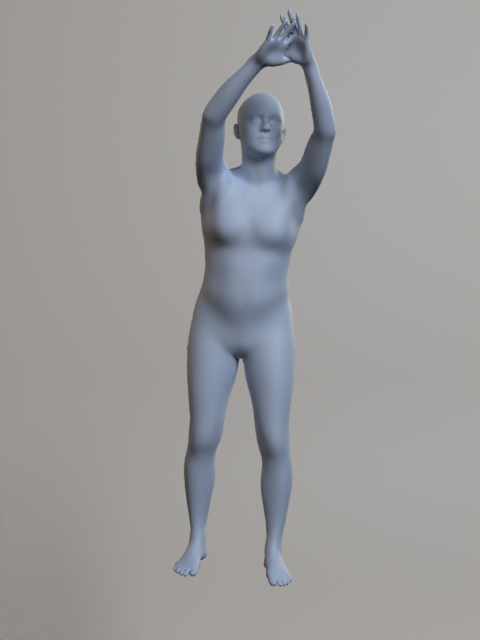} & 
\includegraphics[width=0.11\textwidth]{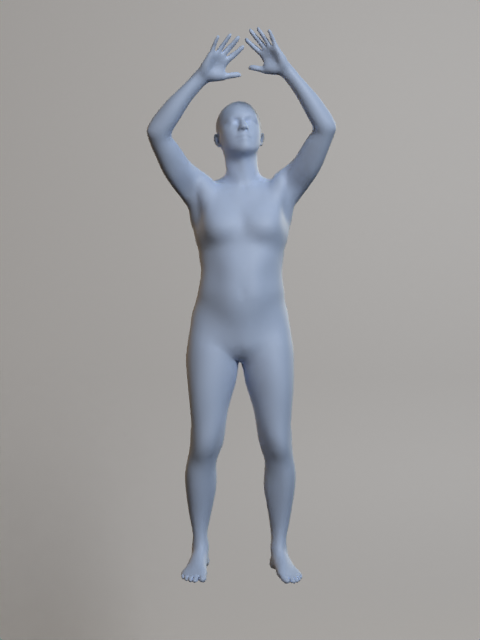} & 
\includegraphics[width=0.11\textwidth]{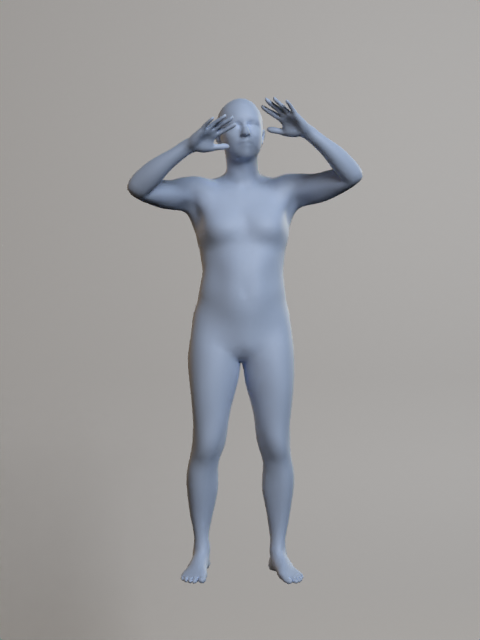} & 
\includegraphics[width=0.11\textwidth]{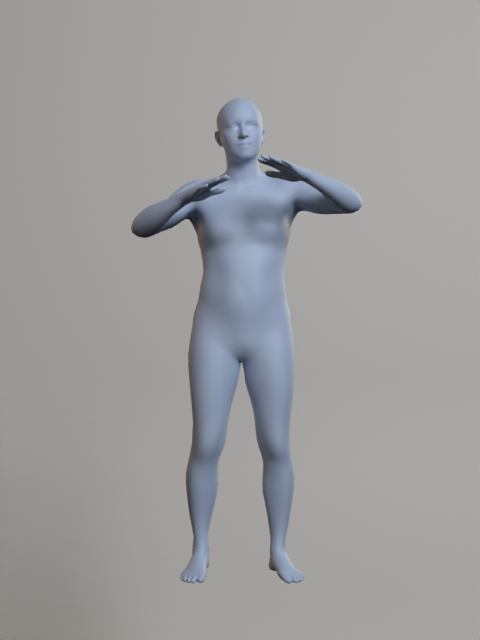} & 
\includegraphics[width=0.11\textwidth]{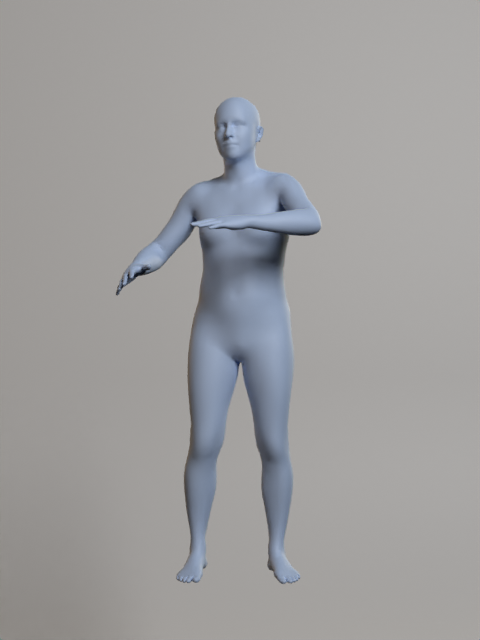} & 
\includegraphics[width=0.11\textwidth]{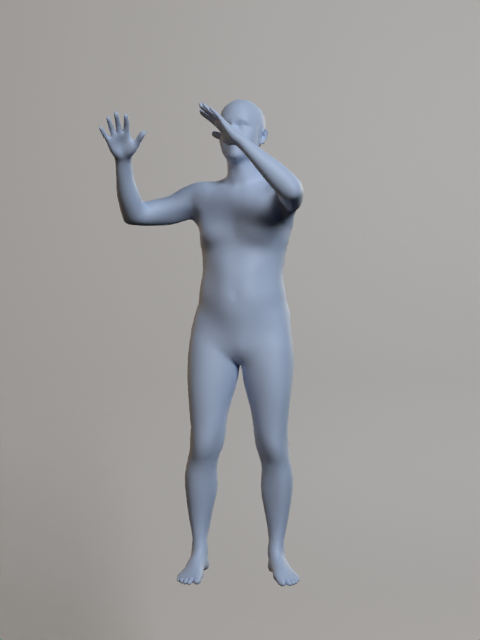} \\ 
\raisebox{23px}{Rec\,} &
\includegraphics[width=0.11\textwidth]{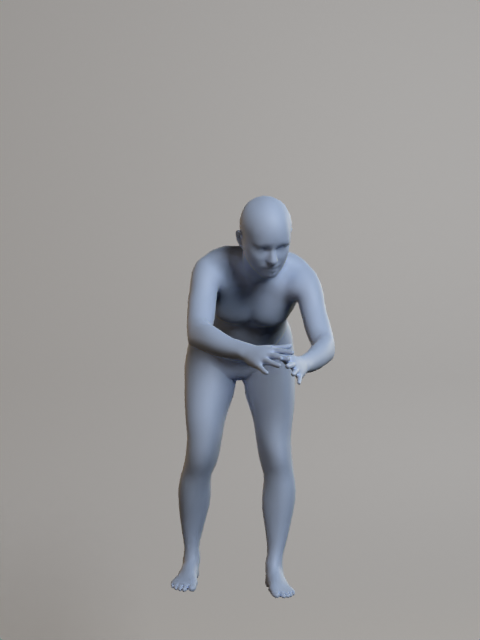} & 
\includegraphics[width=0.11\textwidth]{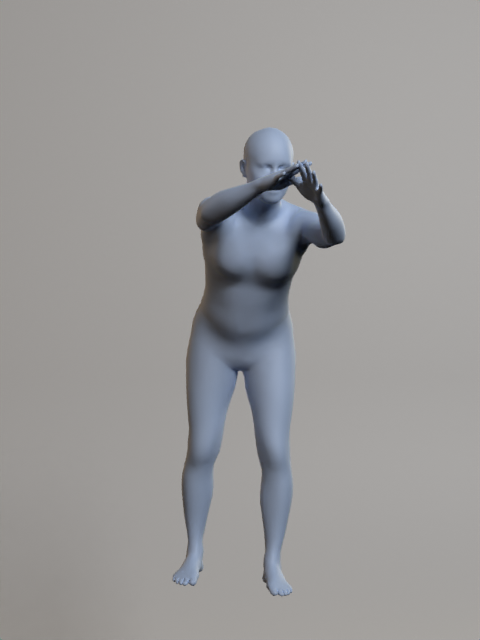} & 
\includegraphics[width=0.11\textwidth]{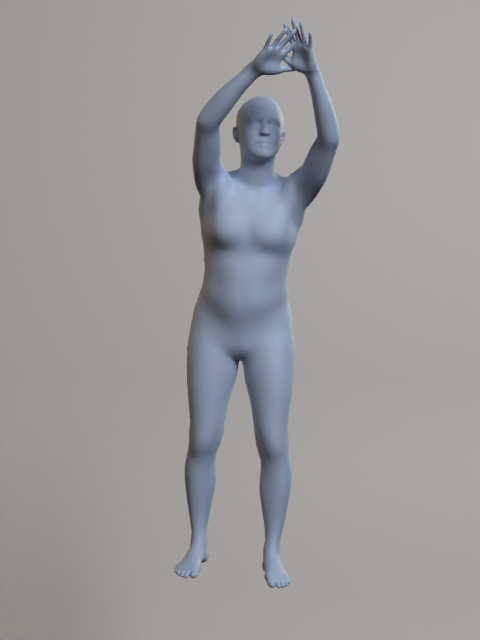} & 
\includegraphics[width=0.11\textwidth]{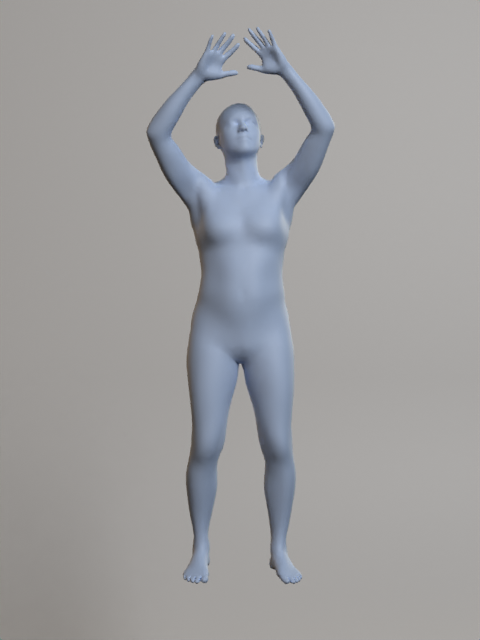} & 
\includegraphics[width=0.11\textwidth]{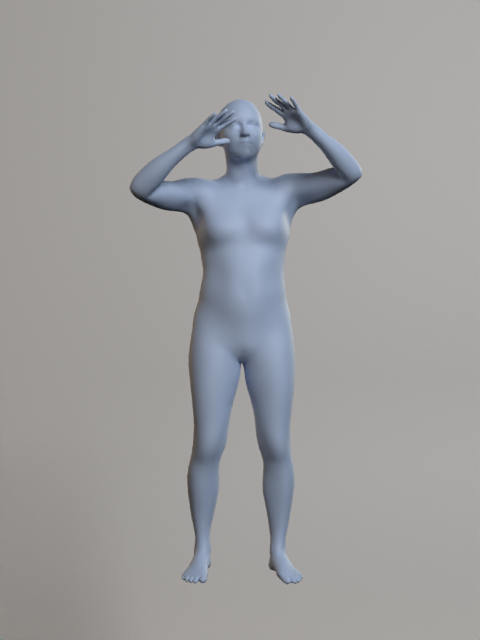} & 
\includegraphics[width=0.11\textwidth]{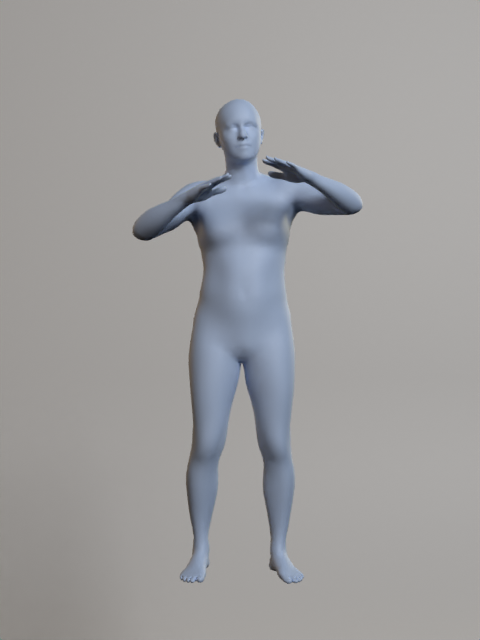} & 
\includegraphics[width=0.11\textwidth]{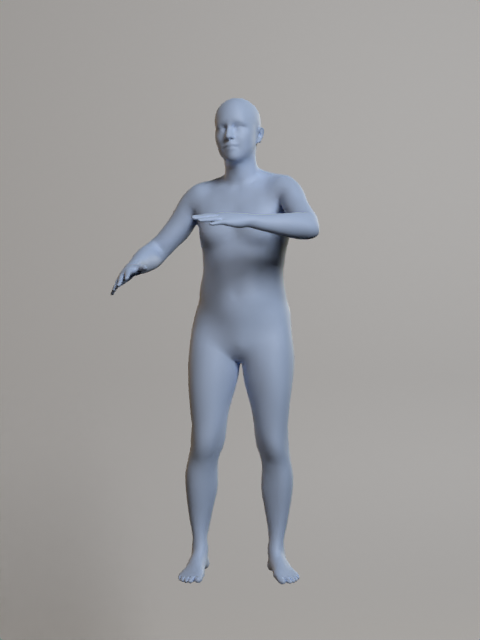} & 
\includegraphics[width=0.11\textwidth]{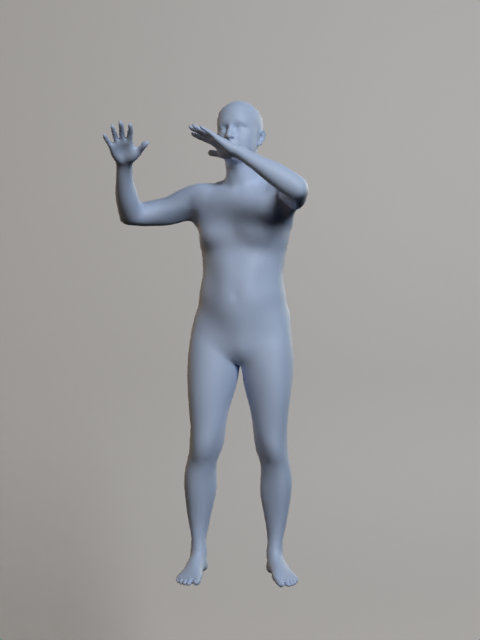} \\ 
\raisebox{23px}{Err\,} &
\includegraphics[width=0.11\textwidth]{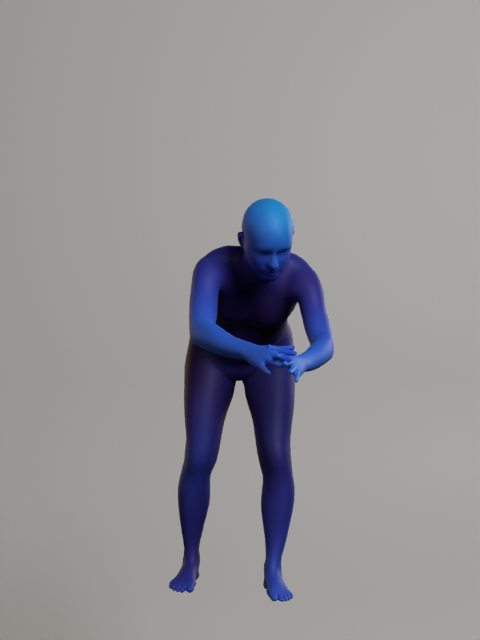} & 
\includegraphics[width=0.11\textwidth]{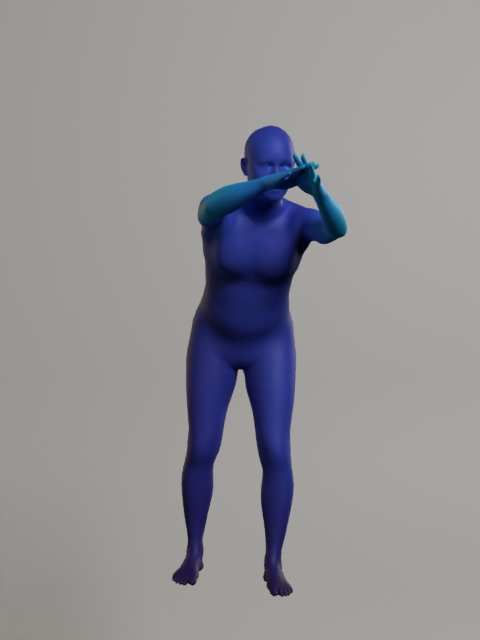} & 
\includegraphics[width=0.11\textwidth]{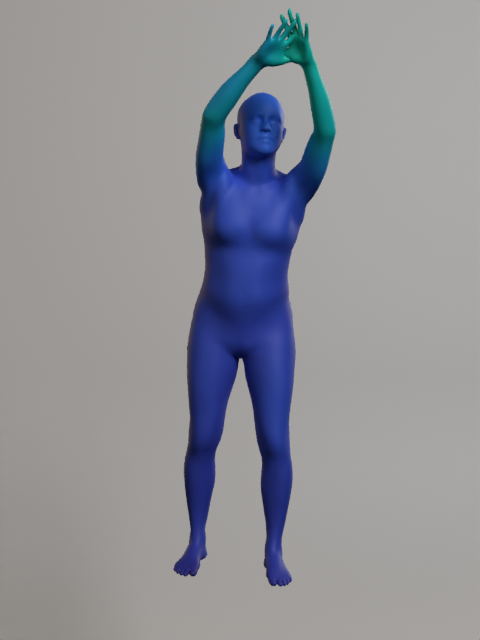} & 
\includegraphics[width=0.11\textwidth]{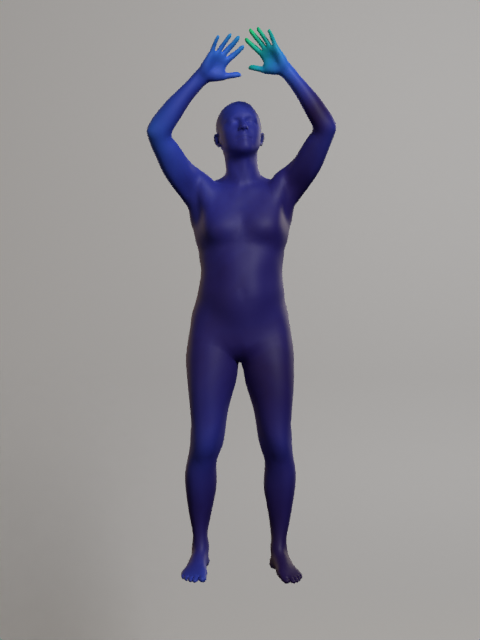} & 
\includegraphics[width=0.11\textwidth]{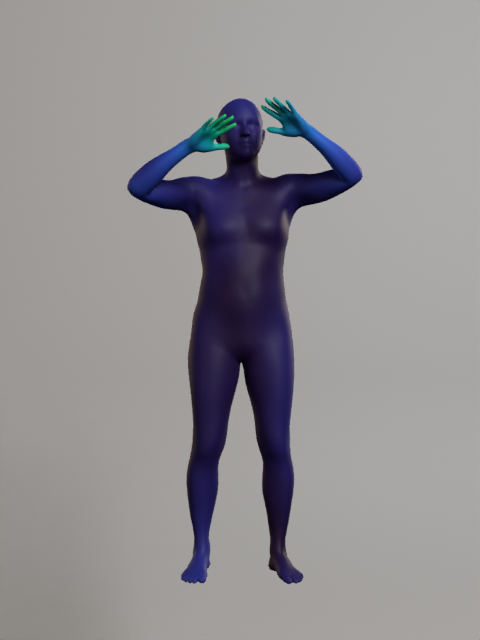} & 
\includegraphics[width=0.11\textwidth]{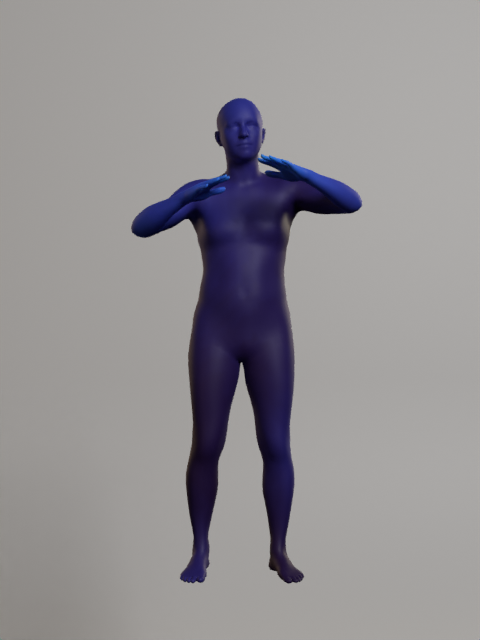} & 
\includegraphics[width=0.11\textwidth]{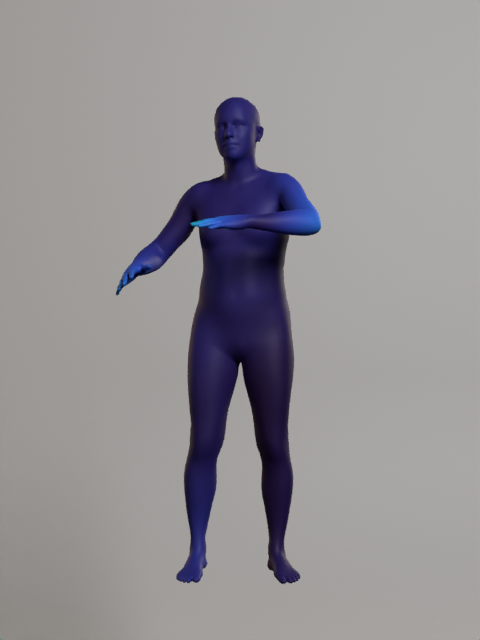} & 
\includegraphics[width=0.11\textwidth]{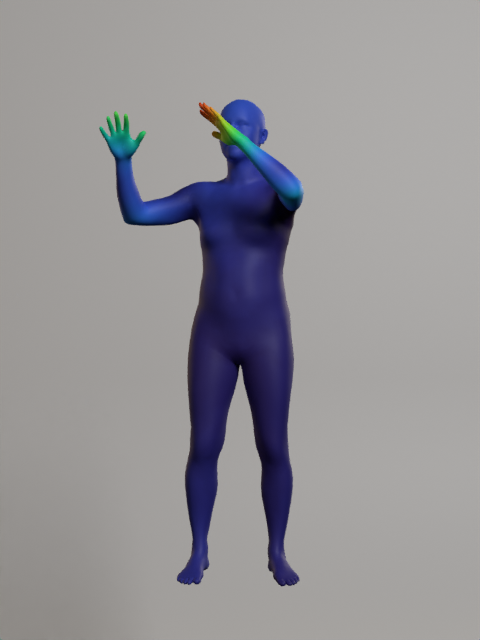} \\ 
& & & & & & & \multicolumn{3}{r}{\scriptsize{0mm} \includegraphics[width=0.10\textwidth]{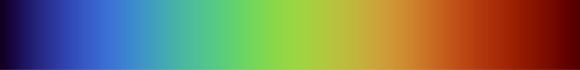} \scriptsize{30mm}}\\
\midrule   \\
\raisebox{23px}{M1\;} &
\includegraphics[width=0.11\textwidth]{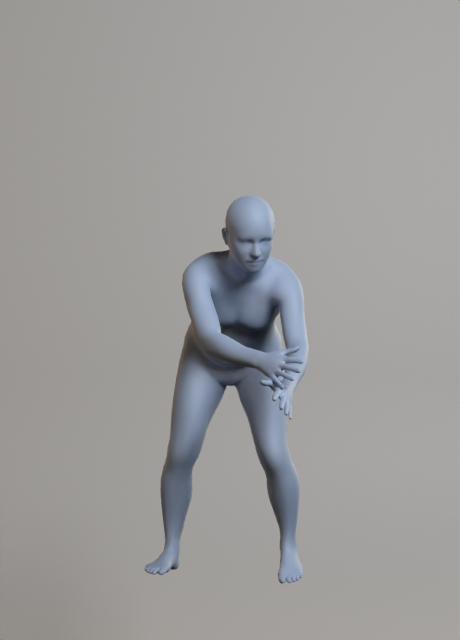} & 
\includegraphics[width=0.11\textwidth]{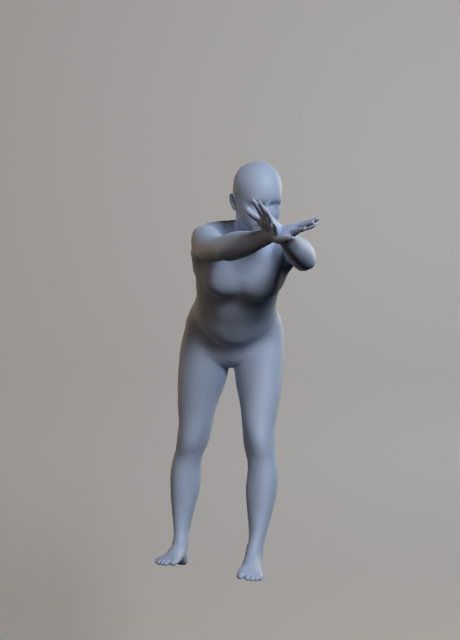} & 
\includegraphics[width=0.11\textwidth]{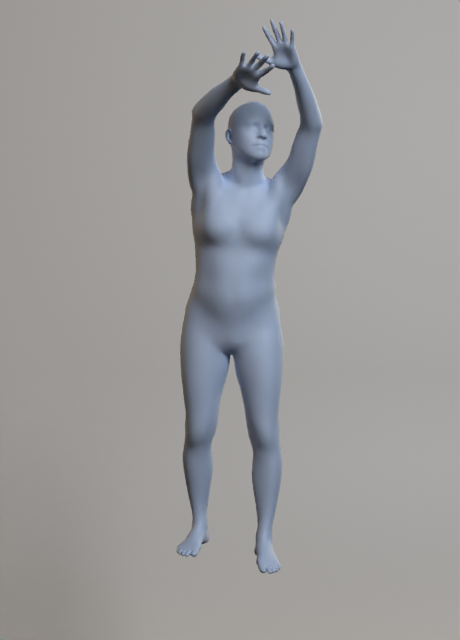} & 
\includegraphics[width=0.11\textwidth]{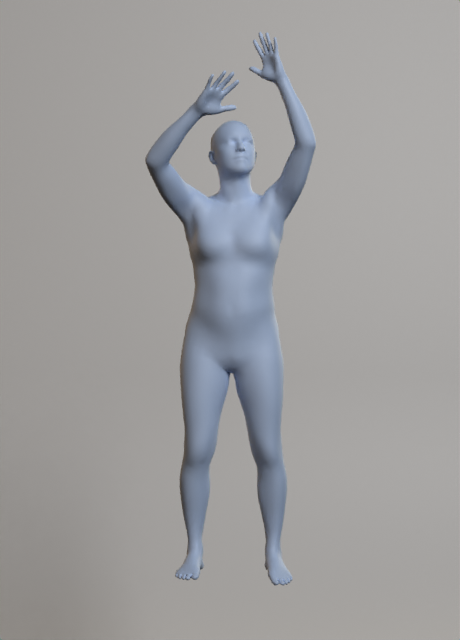} & 
\includegraphics[width=0.11\textwidth]{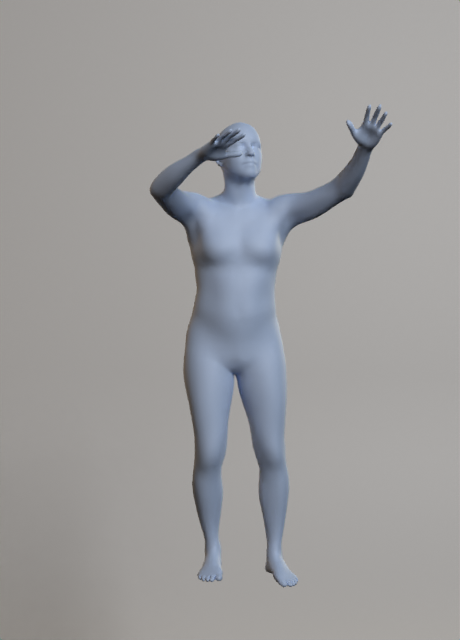} & 
\includegraphics[width=0.11\textwidth]{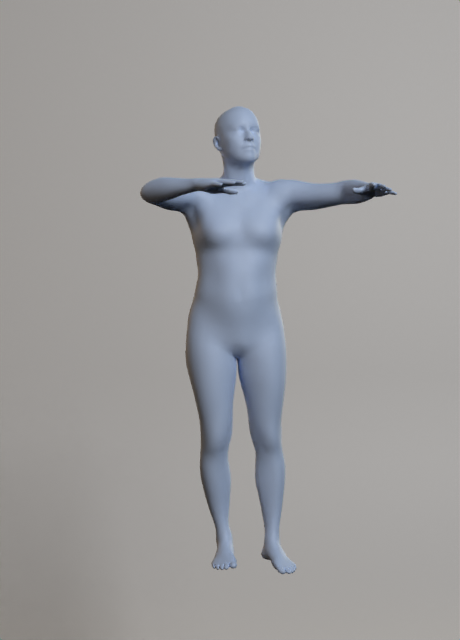} & 
\includegraphics[width=0.11\textwidth]{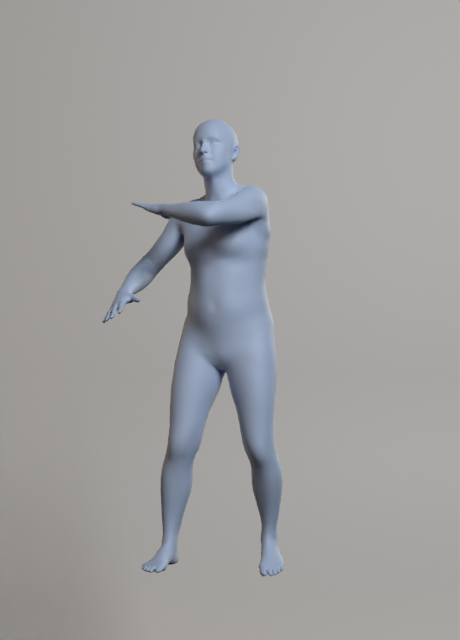} & 
\includegraphics[width=0.11\textwidth]{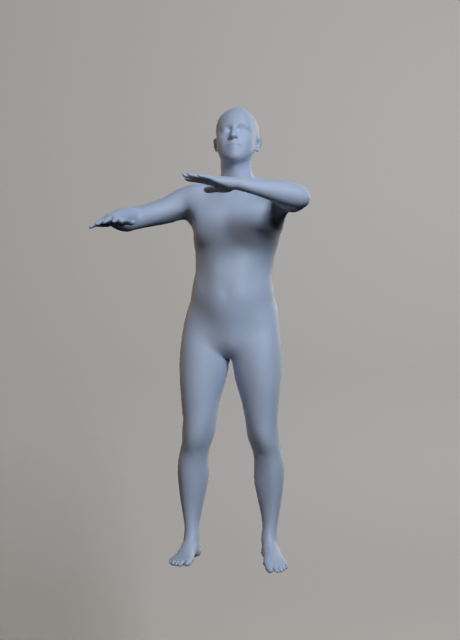} \\ 
\raisebox{23px}{M2\;} &
\includegraphics[width=0.11\textwidth]{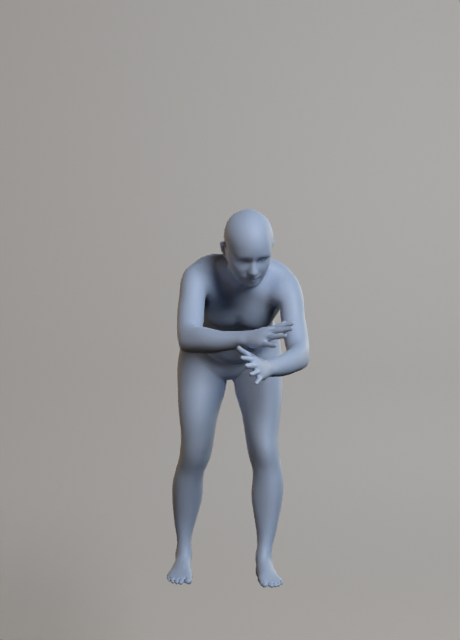} & 
\includegraphics[width=0.11\textwidth]{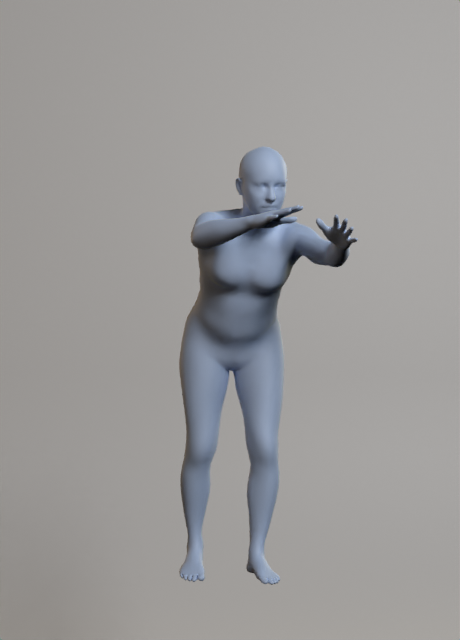} & 
\includegraphics[width=0.11\textwidth]{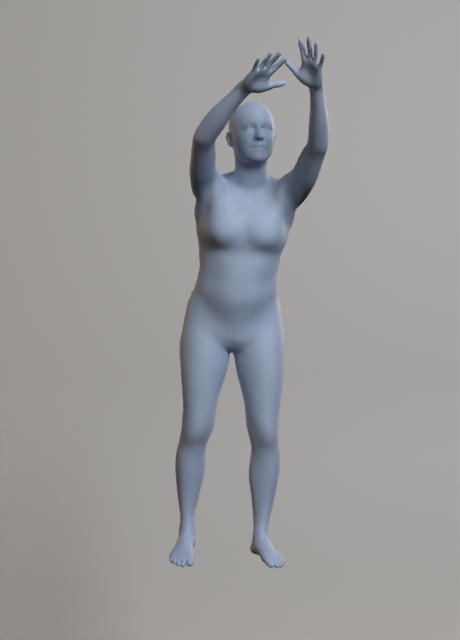} & 
\includegraphics[width=0.11\textwidth]{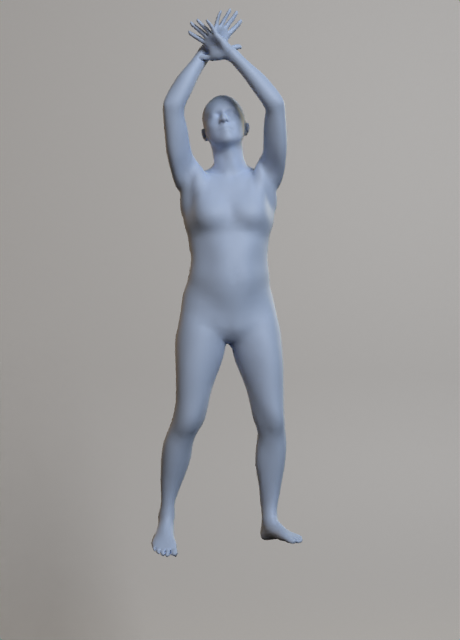} & 
\includegraphics[width=0.11\textwidth]{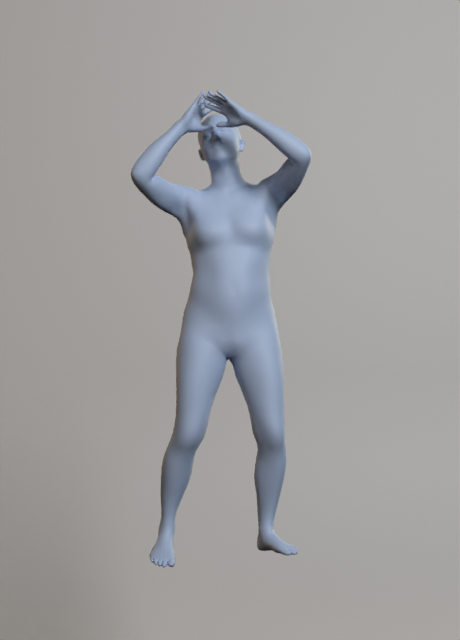} & 
\includegraphics[width=0.11\textwidth]{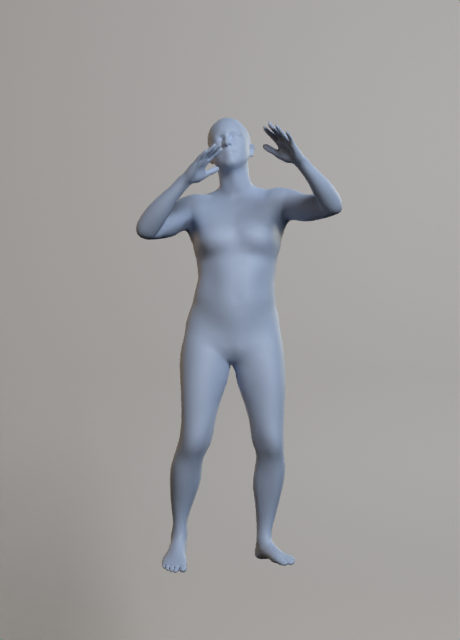} & 
\includegraphics[width=0.11\textwidth]{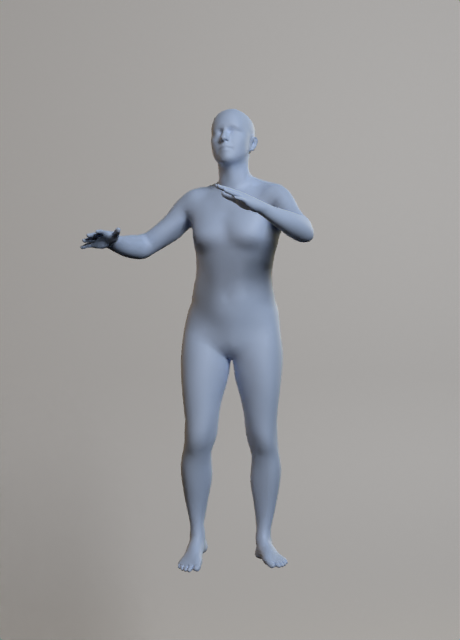} & 
\includegraphics[width=0.11\textwidth]{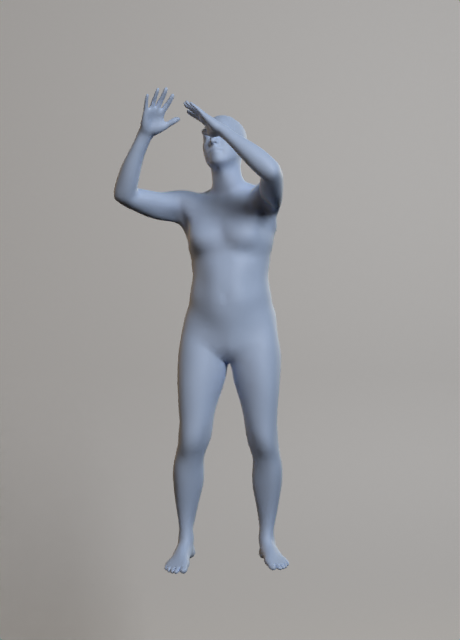} \\ 
\end{tabular}
}
\caption{\textbf{Full Body Reconstruction and Modification.} Upper rows: Reconstruction quality on a motion. Lower rows: Two reconstruction results after modifying the control tokens.}
\label{fig:hmotion_reconstruction}
\end{figure*}

We evaluate our method on the full-body human motion dataset HumanML3D~\cite{guo2022humanML3d}, a combination of the HumanAct12~\cite{guo2020action2motion} and AMASS dataset~\cite{mahmood2019amass}. In \cref{tab:hmotion_ablation}, we report the mean squared reconstruction error of per-frame joint positions measured in degrees. Spline-based Transformers show reconstruction improvements of at least a factor two, reducing the mean joint error of the smallest model (16D) from $\sim\!0.4^\circ$ to $\sim\!0.2^\circ$, and up to $\sim\!0.07^\circ$ for the largest model (64D). We use the joint rotations to compute SMPL body model parameters~\cite{loper2015smpl} and visualize the reconstruction in \cref{fig:hmotion_reconstruction} along with the reconstruction error in mm. Positional encoded transformers have recently received a lot of attention in full-body motion reconstruction and synthesis~\cite{petrovich21actor, tevet2023mdm, duan2021single}. We believe that Spline-based Transformers can help to increase the performance of various state-of-the-art models designed for these applications.

\begin{table}
    \centering
    \captionsetup{justification=centering}
    \caption{\textbf{Human Motion Reconstruction.} Comparison across different latent dimensions. \textbf{Bold} indicates the best overall performance, and \underline{underline} the best in each category. We report the Mean Squared Error (MSE) between joint angles $[deg]$.}
    \label{tab:hmotion_ablation}
        \centering
        \begin{tabular}{
                @{}
                l
                >{\centering\arraybackslash}p{1.1cm}
                >{\centering\arraybackslash}p{1.1cm}
                >{\centering\arraybackslash}p{1.1cm}
                @{}
            }
        \toprule
        {Method} & {16D} & {32D} & {64D}  \\
        \midrule
        ALiBi  & {0.151} & {0.103} & {0.059} \\
        ALiBi-Cat  & {0.153} & {0.103} & {0.051} \\
        Spline (Ours) & {\underline{0.054}} & {\underline{0.022}} & {\textbf{0.006}} \\
        \bottomrule
        \end{tabular}
\end{table}

\noindent \paragraph{\textbf{Motion Editing}} We also show that motions can be modified by applying simple operations to the control points. In \cref{fig:hmotion_reconstruction}, we visualize two resulting motions where the control points have been modified to be closer or further away from the end-points of the spline. The motions preserve the overall style but change in speed and detail. This experiment suggests that  latent splines behave smoothly in a neighborhood and edits result in plausible motions. We observe that motions can easily be toned down or amplified as we show with more results in the accompanying video. The spline-based latent space further allows us to super-sample motions. By sampling the latent spline more densely before decoding, we can achieve up to a 4x upsampling of a motion clip. This method effectively preserves the original motion characteristics, as demonstrated in the video. Multiple splines can be combined to represent longer sequences of motions. Each of the segments can then be modified individually.

\subsection{Geometric Representation}
\label{subsec:strands}

Finally, we show how Spline-based Transformers can also be used to model complex geometry like hair strands~\cite{rosu2022neuralstrands,Zhou_2023}, which present themselves as 3D curves in space. For this experiment, we use a dataset of 343 unique 3D hairstyles~\cite{hairSalon2015}, where each hairstyle contains 10,000 strands, and each strand has 100 points. We are interested in representing only the strand geometry in our experiment, so we consider the strands across the hairstyles as individual 3D curves in space. We normalize the root position of each strand by translating it to the origin. Each normalized strand is therefore a sequence of a 100 vertices and is used to train a transformer autoencoder as before with an L2 reconstructive loss. We report the reconstruction error of strands from 10 test hairstyles in~\cref{tab:strand_ablation}. While a thorough comparison to state-of-art methods~\cite{rosu2022neuralstrands,Zhou_2023} is required to demonstrate the real effectiveness of Spline-based Transformers for this task, our initial tests indicate that they could be an interesting architectural alternative. For the purpose of visualizing the reconstructed strands as a coherent hairstyle, we apply the ground truth root position to the reconstructed strands in \cref{tab:strand_ablation}.

\begin{table}[h]
    \centering
    \captionsetup{justification=centering}
    \caption{\textbf{Strand Reconstruction.} Comparison across different latent dimensions. \textbf{Bold} indicates the best overall performance, and \underline{underline} the best in each category. Performance is measured in Mean Squared Error (MSE).}
    \label{tab:strand_ablation}

    \begin{minipage}{.48\linewidth}
        \centering
        \begin{tabular}{lccc}
        \hline
        Method & 8D & 16D & 32D \\ 
        \hline
        ALiBi & 4.8e-3 & 2.0e-3 & 1.5e-3  \\
        ALiBi-Cat & 4.6e-3 & 1.9e-3 & 1.2e-3 \\
        Spline (Ours) & \underline{1.09e-3} & \underline{1.06e-3} & \textbf{9.4e-4} \\
        \hline
        \end{tabular}
        \end{minipage}%
    \begin{minipage}{.48\linewidth}
        \centering
        \includegraphics[width=\linewidth]{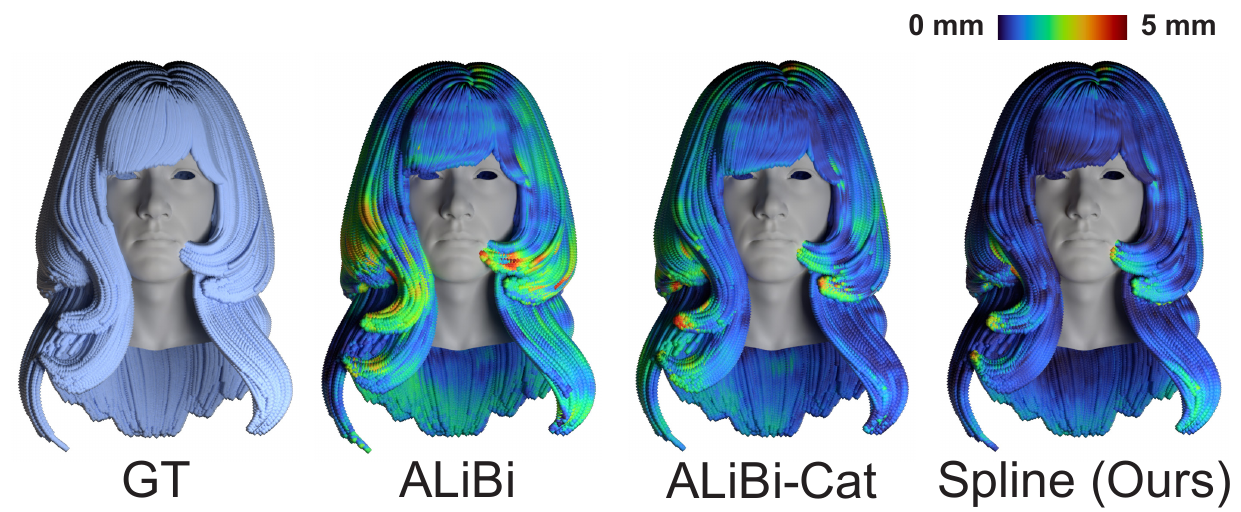}
    \end{minipage}
\end{table}

\subsection{Practical Limitations} 
\label{subsec:training}

\begin{figure}
    \centering
    \begin{subfigure}{0.45\textwidth}
        \centering
        \includegraphics[width=0.75\textwidth]{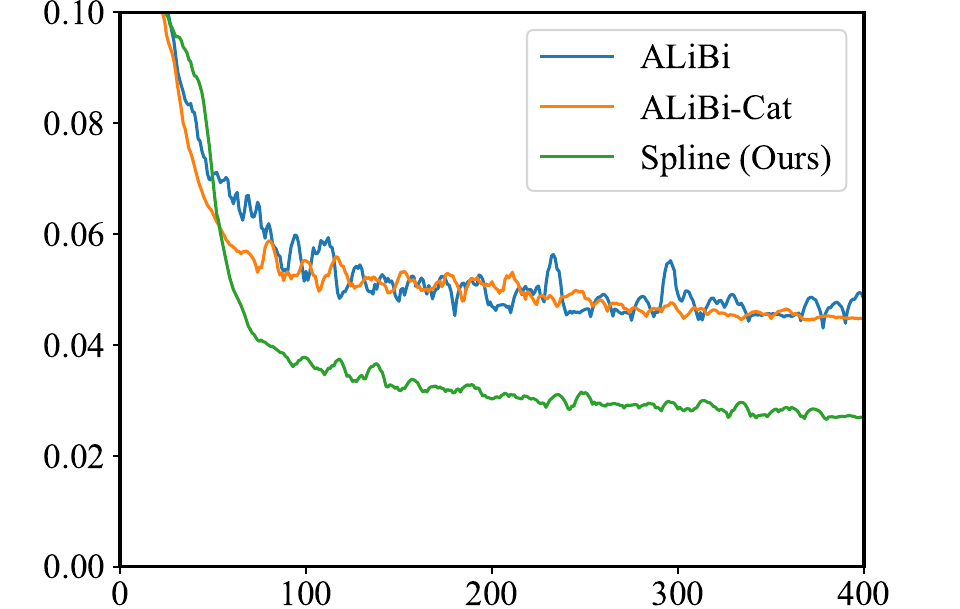}
        \caption{validation loss}
        \label{subfig:val_loss}
    \end{subfigure}
    \hfill
    \begin{subfigure}{0.45\textwidth}
        \centering
        \includegraphics[width=0.75\textwidth]{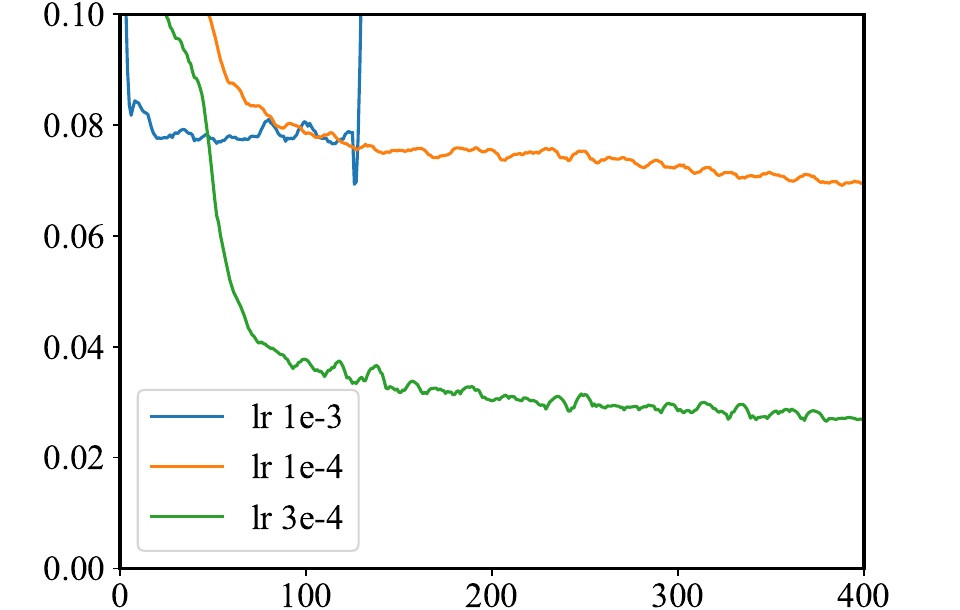}
        \caption{learning rate sensitivity}
        \label{subfig:lr_sensitivity}
    \end{subfigure}
    \caption{\textbf{Training Performance (AFHQ).} (a) shows the validation loss of the different methods. (b) shows the sensitivity of Spline-based Transformers to the learning rate. }
    \label{fig:training}
\end{figure}

Spline-based Transformers not only achieve a better performance than conventional transformers as demonstrated by our experiments, but can also achieve this performance improvement much faster than the traditional positional encoded models. \cref{subfig:val_loss} shows the validation loss on an image experiment. 

While they converge faster, we observe that the Spline-based Transformers are sensitive to learning rates. \cref{subfig:lr_sensitivity} shows the same run with three different learning rates. A large learning rate can lead to a model collapse where the control points start to converge to the same point in latent space; leading to the same latent code for each token in the input sequence and the model not being able to recover from this state. We also observe that having a too small learning rate can harm performance significantly. We believe that a specialized scheduling strategy could significantly improve the stability and performance of Spline-based Transformers and leave this as future work.

\begin{figure}
    \centering
    \begin{subfigure}{0.45\textwidth}
        \centering
        \includegraphics[width=0.75\textwidth]{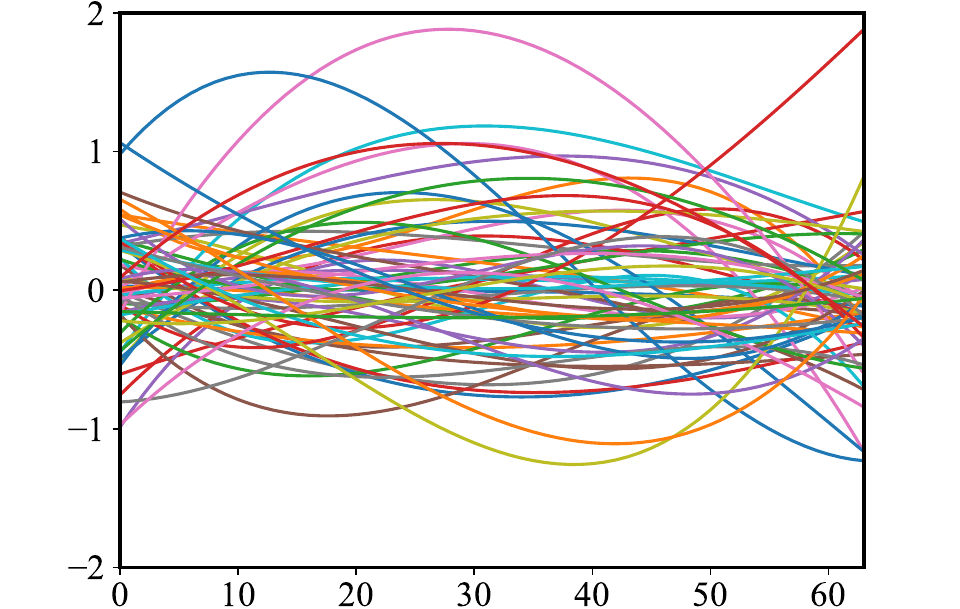}
    \end{subfigure}
    \hfill
    \begin{subfigure}{0.45\textwidth}
        \centering
        \includegraphics[width=0.75\textwidth]{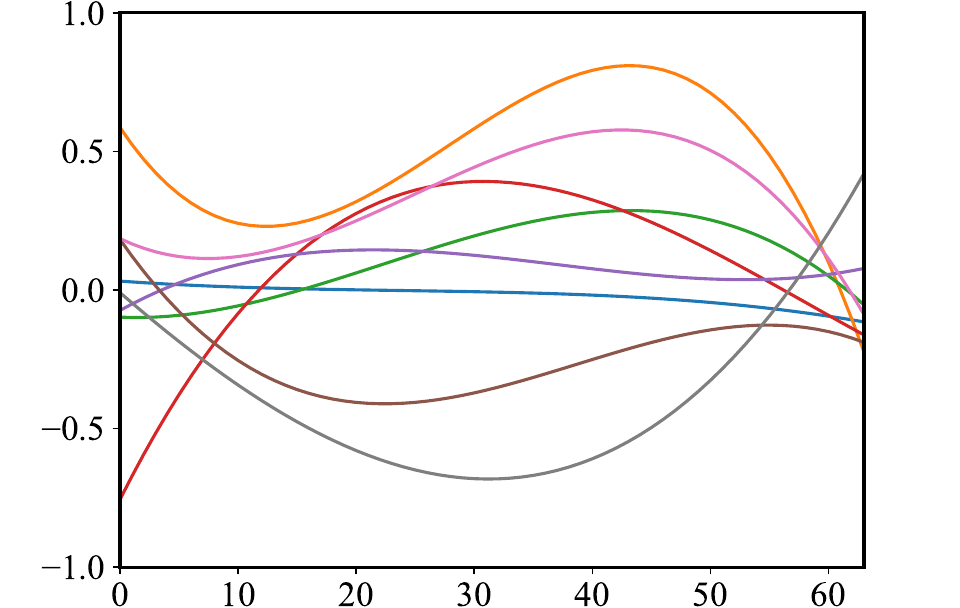}
    \end{subfigure}
    \caption{\textbf{Latent Splines (AFHQ).} Visualization of predicted latent spline trajectories in $d=64$ and $d=8$. $x$-axis: position along the spline; $y$-axis: value of the feature.}
    \label{fig:latentsplines}
\end{figure}

\subsection{Latent Space Visualization}
After training, we can visualize the latent spline trajectories (see \cref{fig:latentsplines}). The trajectories show similar characteristics with sinusoidals but are more complex and asymmetric curves. In some parts, the change of the features is more rapid, while other dimensions propagate the same value over the whole sequence length.

\section{Conclusion}

In this work, we introduced Spline-based Transformers, a new class of Transformer models that eliminate the need for absolute positional encoding by combining temporal and contextual information into a single trajectory, represented by a latent spline curve. We presented the superior performance of Spline-based Transformers across a variety of datasets, from simple curves to complex animation data and images. The experiments show significant performance improvements over traditional positional encoded transformer models. Spline-based Transformers are trivial to implement yet effective and have no additional computational overhead. We identify improvements to the training stability as future work to reduce the sensitivity to training hyperparameters such as, \eg, the learning rate. The spline-based latent space introduced by our method opens up a new way to interact with latent spaces using straightforward modifications of the latent control points. We hope that our work encourages research towards a new class of transformers with controllable latent spaces across a variety of applications.

%
%
\bibliographystyle{splncs04}
\bibliography{egbib}

\newpage

\title{Spline-based Transformers \break Supplementary Material} 


\author{
    Prashanth Chandran\inst{1}\textsuperscript{*}
    \and
    Agon Serifi\inst{2,3}\textsuperscript{*}
    \and \newline
    Markus Gross\inst{1,3}
    \and
    Moritz B{\"a}cher\inst{2}
}

\authorrunning{P.~Chandran and A.~Serifi et al.}

\institute{
    DisneyResearch|Studios, Switzerland 
    \and
    Disney Research, Switzerland \\ 
    \email{\{prashanth.chandran, moritz.baecher\}@disneyresearch.com} 
    \and
    ETH Zurich, Switzerland \\
    \email{\{agon.serifi, grossm\}@inf.ethz.ch}
}

\maketitle

\footnotetext{\textsuperscript{*} equal contribution.}

\begin{figure}
   \centering
   \includegraphics[width=\linewidth]{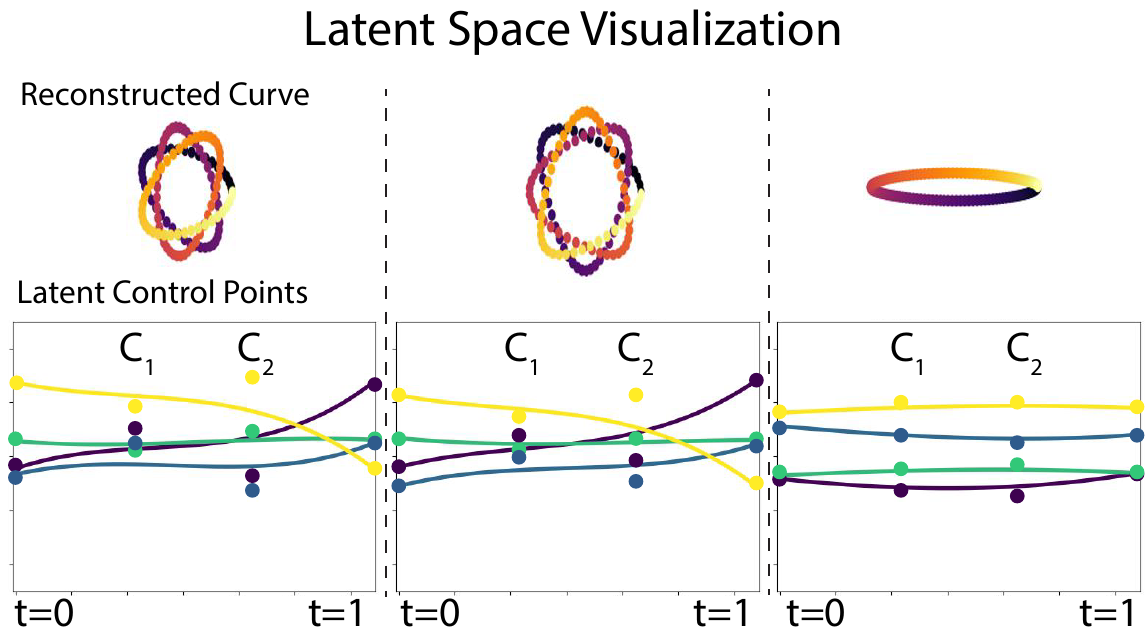}
    \caption{We show Hypotrochoid curves generated by our method along with their corresponding latent control points (shown as dots) and the latent trajectories obtained using a cubic B\'ezier interpolator. Curves that look similar in Cartesian space (Cols 1, 2) seem to have similar latent controls and trajectories, while smoother curves (Col 3) seem to have smoother latent trajectories.}
    \label{fig:trajectory}
\end{figure}

\section{More About Control Points}
\noindent \textbf{Role In Downstream Applications} In our work, we evaluate Spline-based Transformers mainly from the point of view of reconstruction tasks as they are a common proxy task for pre-training feature backbones for various downstream applications (such as classification, segmentation, etc). Even when trained to reconstruct input sequences, Spline-based Transformers already seem to capture the similarities and differences between the input sequences in their latent control points. In \cref{fig:trajectory}, we show the predicted control points of two similar-looking curves (Cols 1, 2) alongside a different curve (Col 3)  with their corresponding latent trajectories in the second row. Complex curves appear to have more nonlinear latent trajectories (Cols 1, 2), while the smoother curve (Col 3) has a more linear latent trajectory. While further experimentation is certainly required, our early results indicate that Spline-based Transformers will find use in various other applications, including representation learning, classification \etc. 

\begin{figure}
   \centering
   \includegraphics[width=\linewidth]{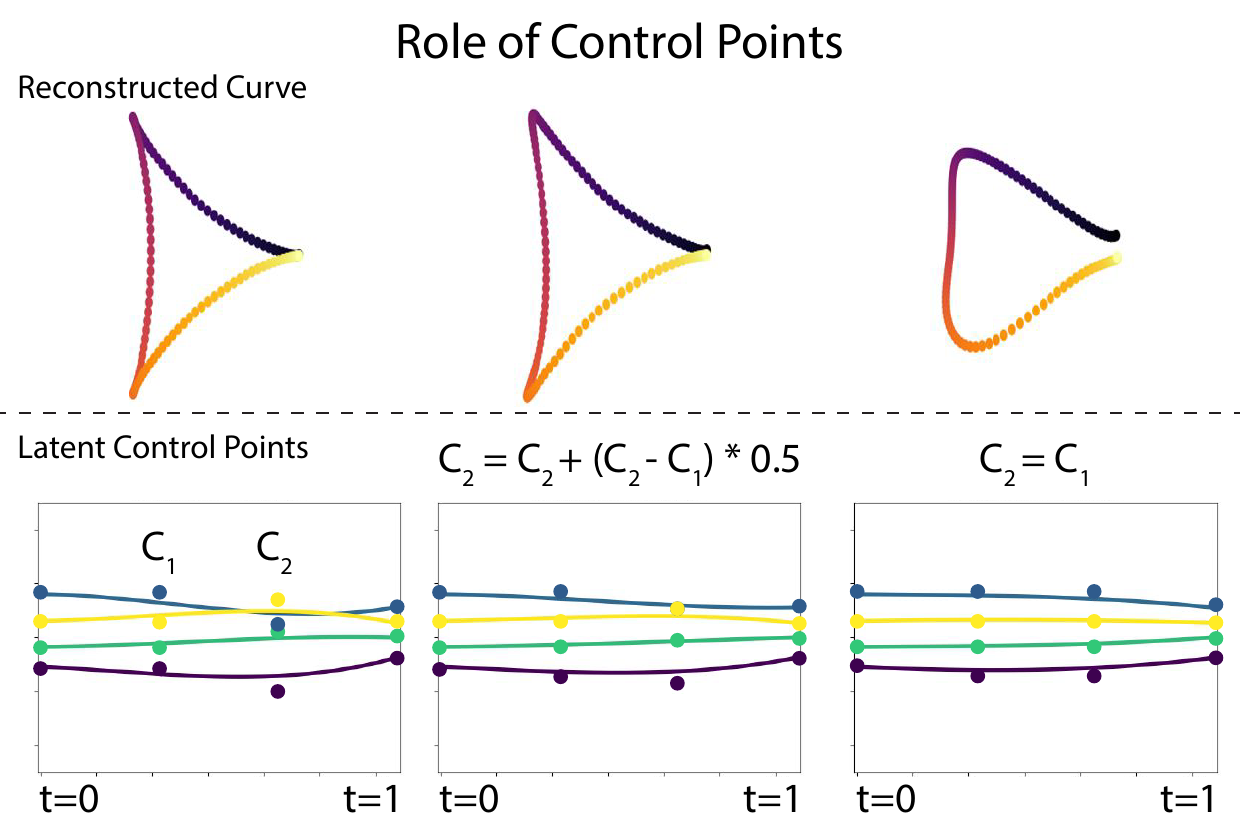}
    \caption{Modifying the control points alters the latent trajectories according to the chosen B-Spline (cubic B\'ezier in our case). Here we incrementally modify control point $C_{2}$ to be closer to $C_{1}$ and visualize the change in the latent space. The corresponding reconstructed curve is shown in the top row.}
    \label{fig:interpolation}
\end{figure}

\noindent \textbf{Latent Control Manipulation}
The number of control tokens/points depends on the type of B-Spline (cardinal B-Splines, cubic B\'ezier, etc.) used. So far we have used cubic B\'ezier splines, which are defined by four control tokens across all data modalities. These control points determine the trajectory of the input sequence in the latent space of a Spline-based Transformer. For cubic B\'ezier splines, the first and the last latent control points determine the start and end of the latent trajectory, while the second $C_{1}$ and third $C_{2}$ control points define the shape of the latent spline. \cref{fig:interpolation} demonstrates the effect of modifying the control point $C_{2}$ on 2D Hypotrochoids and the effect this has on reconstruction. 

\section{Implementation Details}

In the following, we present the implementation details for our Spline-based Transformer to help with reproducibility. In \cref{tab:params}, we enumerate the various hyperparameters used in building the Spline-based Transformer for each of our experiments. 

\begin{algorithm}[H]
\SetKwInput{KwInput}{Input}                
\SetKwInput{KwOutput}{Output}              
\SetKwInput{KwParameter}{Param}       
\SetAlgoLined
\KwInput{Sequence $x$ of dimensions $(seq\_len \times data\_dim)$. \\}
\KwParameter{Number of control points $controls$. Dimension of spline $d$. Network layers \{encoder,decoder,T5\}\_layers.}
\KwOutput{Predicted sequence $\tilde{x}$.}
\BlankLine

{\color{blue} \tcc{components}}

MLPEnc = Sequential(encoder\_layers)

MLPDec = Sequential(decoder\_layers)

TEnc = TransformerEncoder(T5\_layers)

TDec = TransformerDecoder(T5\_layers)

$\mathbf{p} =$ ParameterList([Parameter($d$) for \_ in range($controls$)])

{\color{blue} \tcc{forward pass}}
$feat\_in =$ MLPEnc$(x)$ \\
$enc\_inp =$ cat($*\mathbf{p}$, $feat\_in$) {\color{black}\Comment{concat control points}}\\
$enc\_out =$ TEnc($enc\_inp$) \\
\BlankLine
\BlankLine
$c\_points =$ $enc\_out$[:controls] \\
$latents =$ \textbf{EVALUATE}($c\_points$, $seq\_len$)  {\color{black}\Comment{\cref{eq:spline}}}\\
\BlankLine
\BlankLine
$feat\_out =$ TDec($latents$) \\
$\tilde{x} =$ MLPDec($feat\_out$) \\

\caption{Spline-based Transformer}
\label{algo:main}
\end{algorithm}

\begin{algorithm}[H]
\DontPrintSemicolon
\SetKwInput{KwInput}{Input}                
\SetKwInput{KwOutput}{Output}              
\SetAlgoLined
\KwInput{Four control points $c\_points$ and sequence length $seq\_len$. \\}
\KwOutput{Uniformly evaluated latent spline $\mathbf{s}$.}
\BlankLine

t = linspace(0, 1, $seq\_len$)

$s = ((1 - t)^3 \cdot c\_points[0]$ \\
$\qquad +\, 3.0 \cdot (1 - t)^2 \cdot t \cdot c\_points[1]$ \\
$\qquad +\, 3.0 \cdot (1 - t) \cdot t^2 \cdot c\_points[2]$ \\
$\qquad +\, t^3 \cdot c\_points[3]$) \\

\Return s
\caption{\textbf{EVALUATE} CubicBézier}
\label{algo:bezier}
\end{algorithm}

Spline-based Transformers only require simple modifications to a traditional transformer model and easily fit into various existing setups. Alg.~\ref{algo:main} presents a general implementation of Spline-based Transformers using a torch-like syntax. Different layers can be chosen for MLP encoder/decoder, \eg, in the case of image data, 2D convolution layers can be used. Similarly, different transformer blocks can be stacked together to build our transformer encoder/decoder. Our control tokens $\mathbf{p}$ are learnable parameters (Line 6) and are initialized from a normal distribution. These learnable control tokens are concatenated to the sequence of tokens after the MLP encoder (Line 9) to result in the transformer encoder's input sequence $enc\_inp$. The tokens corresponding to the control tokens are interpreted as latent control points at the output of the encoder, which are then evaluated as a spline (Line 11-12). In our experiments, we used cubic Bézier splines (with four control points), and Alg.~\ref{algo:bezier} shows how we evaluate them to obtain a latent token sequence for the decoder. Finally, the interpolated latents are passed through the transformer decoder and the shared MLP decoder and ultimately mapped back into their original space. An L2 loss function between the input and output sequence is used to train our system end-to-end.

\begin{table}
\centering
\caption{Parameters and Hyperparameters.}
\label{tab:params}
\footnotesize
\begin{tabularx}{\linewidth}{
  L{\dimexpr.05\linewidth-2\tabcolsep} 
  L{\dimexpr.24\linewidth-2\tabcolsep} 
  C{\dimexpr.1\linewidth-2\tabcolsep} 
  C{\dimexpr.076\linewidth-2\tabcolsep}  
  C{\dimexpr.076\linewidth-2\tabcolsep} 
  C{\dimexpr.076\linewidth-2\tabcolsep} 
  C{\dimexpr.076\linewidth-2\tabcolsep} 
  C{\dimexpr.076\linewidth-2\tabcolsep} 
  C{\dimexpr.076\linewidth-2\tabcolsep} 
  C{\dimexpr.076\linewidth-2\tabcolsep} 
  C{\dimexpr.076\linewidth-2\tabcolsep} 
}
\toprule
Sec & Experiment & Param. & $d$ & $n$ & $h$ & $c$ & FFN & BS & lr & PS \\
\midrule
\multirow{4}{*}{\ref{subsec:synthetic}} & Lissajous (3D) & 0.20M & 3 & 4 & 4 & 64 & 1 & 256 & $1e^{-3}$ & - \\
& Hypotrochoids (4D) & 0.20M & 4 & 4 & 4 & 64 & 1 & 256 & $1e^{-3}$ & - \\
& Bézier (2D) & 0.20M & 2 & 4 & 4 & 64 & 1 & 1024 & $1e^{-3}$ & - \\
& Bézier (64D) & 0.43M & 64 & 4 & 4 & 128 & 1 & 1024 & $1e^{-3}$ & - \\
\midrule
\multirow{3}{*}{\ref{subsec:images}} & CIFAR10 & 0.85M & $2^{5,6,7}$ & 4 & 8 & 128 & 1 & 512 & $3e^{-4}$ & 4 \\
& AFHQ & 1.65M & $2^{5,6,7}$ & 4 & 8 & 128 & 1 & 512 & $3e^{-4}$ & 32 \\
& Face images & 1.65M & $2^{5,6,7}$ & 4 & 8 & 128 & 1 & 512 & $3e^{-4}$ & 32 \\
\midrule
\multirow{2}{*}{\ref{subsec:animation}} & Faces & 12.3M & $2^{5,6,7,8}$ & 4 & 8 & 256 & 1 & 32 & $5e^{-5}$ & - \\
& Motions & 0.63M & $2^{4,5,6}$ & 4 & 8 & 128 & 1 & 1024 & $3e^{-4}$ & - \\
\midrule
\multirow{1}{*}{\ref{subsec:strands}} & Strands & 0.21M & $2^{3,4,5}$ & 4 & 8 & 64 & 1 & 128 & $1e^{-3}$ & - \\
\bottomrule
\end{tabularx}
\end{table}

\cref{tab:params} summarizes the parameters and hyperparameters used to conduct the experiments presented in the paper. The latent dimension $d$, number of stacked transformer layers $n$, number of transformer heads per layer $h$, and the feature size of each layer $c$. Internally, the transformer layers consist of Feed-Forward Networks (FFN), two fully connected layers with non-linear activation GELU. FFNs can have an inner dimension that is 1-4x larger than their outer dimension. We keep the inner structure equal and set the factor to 1x. BS is the batch size, and lr is the learning rate. For the image results, we used a patch size PS. The number of parameters corresponds to the model with the largest latent dimension.

\section{Additional Results}

\begin{figure*}[hb]
    \centering
    \includegraphics[width=\textwidth]{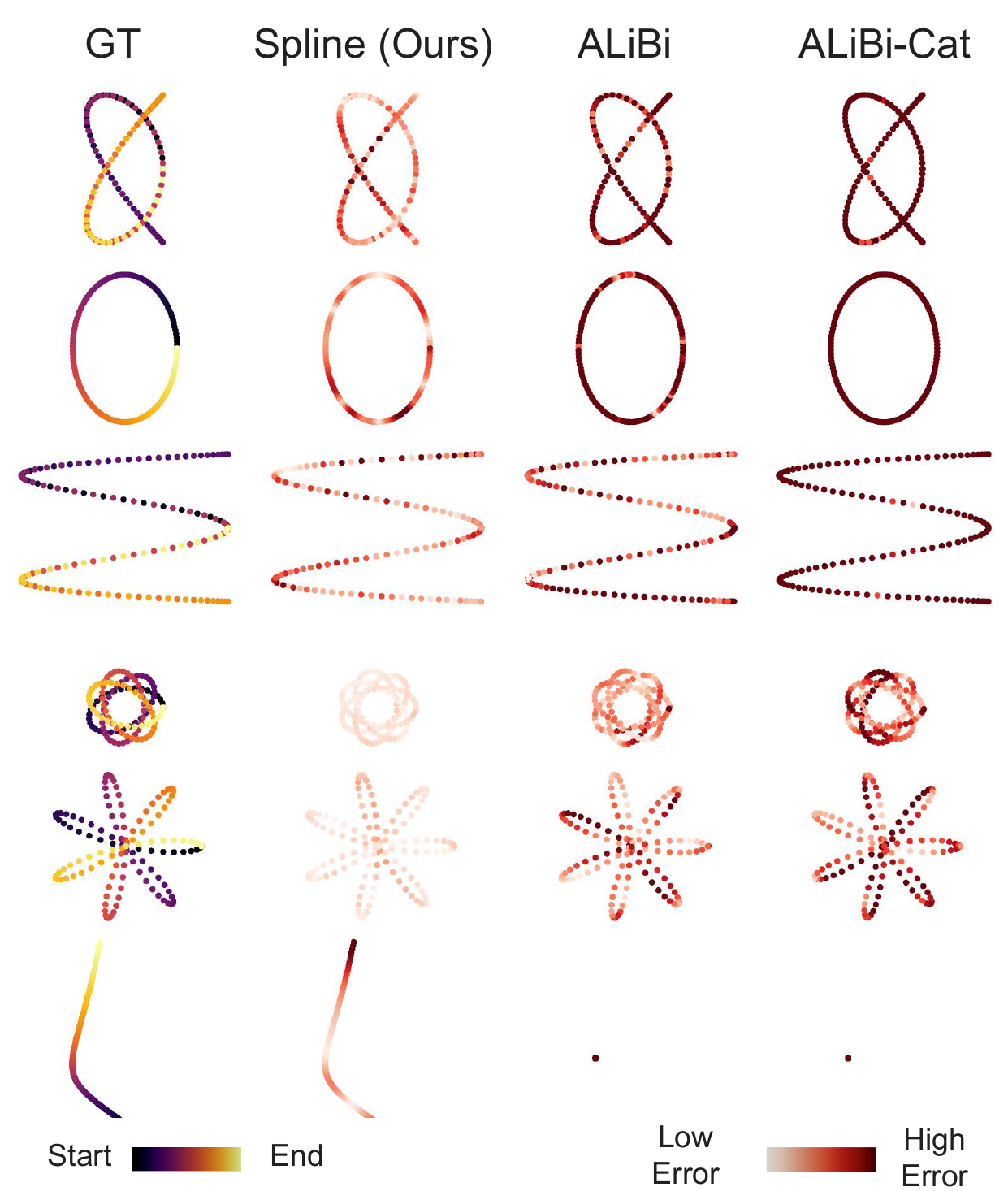}
    \caption{Additional reconstruction results for 2D curves.}
    \label{fig:curves}
\end{figure*}

\begin{figure*}[hb]
\centering
\includegraphics[width=0.4\textwidth]{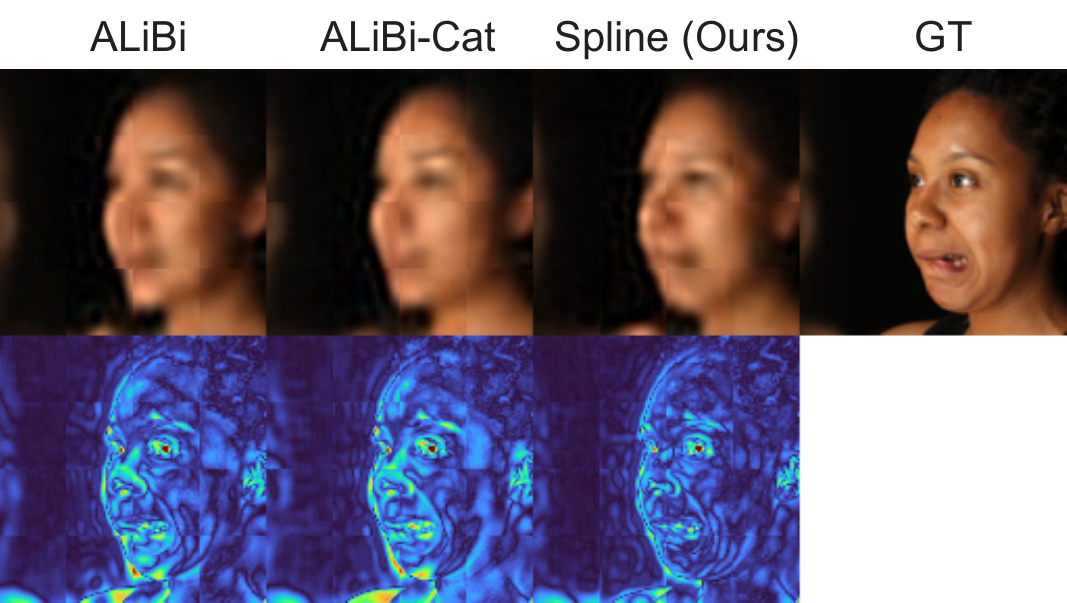}
\includegraphics[width=0.4\textwidth]{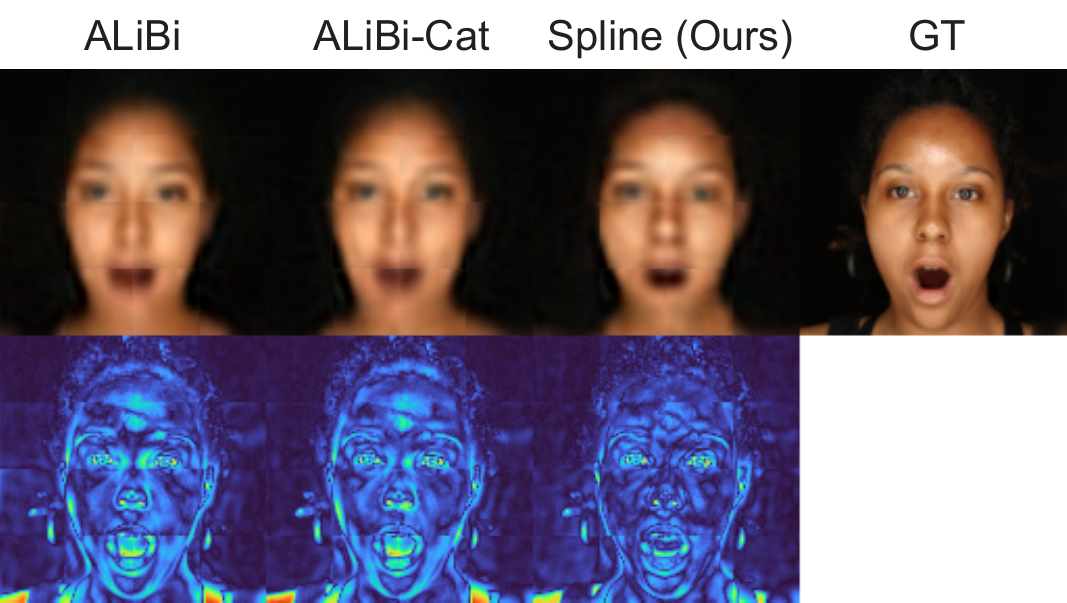}
\includegraphics[width=0.4\textwidth]{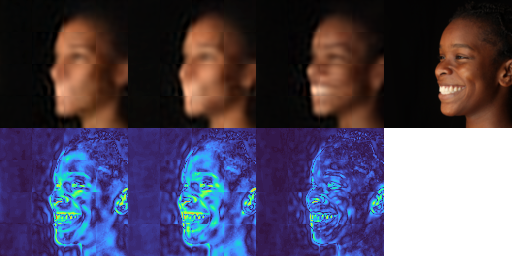}
\includegraphics[width=0.4\textwidth]{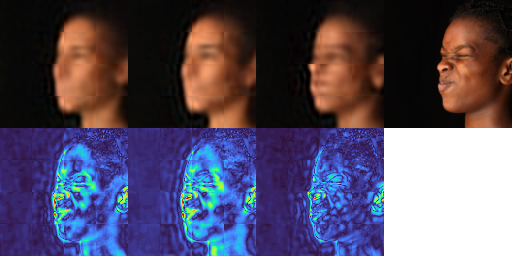}
\includegraphics[width=0.4\textwidth]{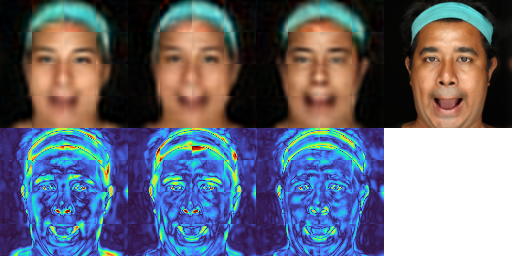}
\includegraphics[width=0.4\textwidth]{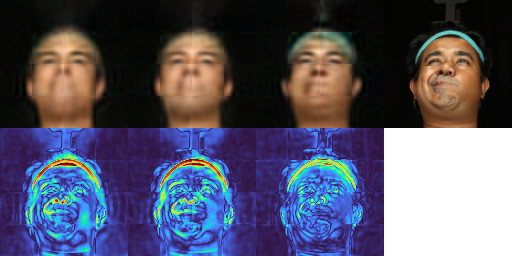}
\includegraphics[width=0.4\textwidth]{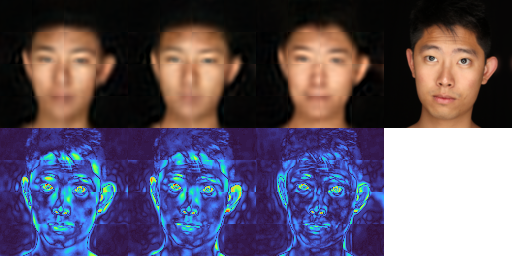}
\includegraphics[width=0.4\textwidth]{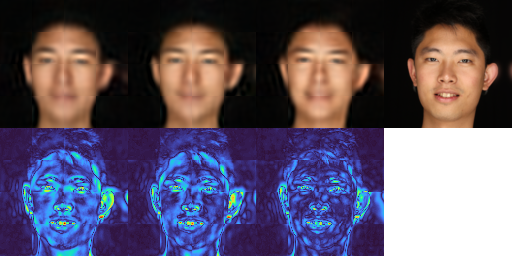}
\includegraphics[width=0.4\textwidth]{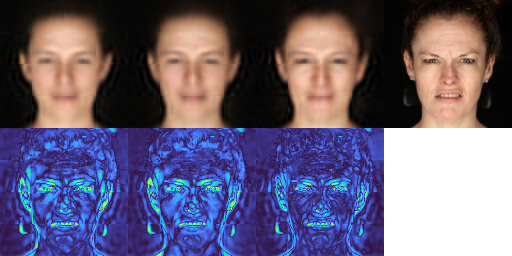}
\includegraphics[width=0.4\textwidth]{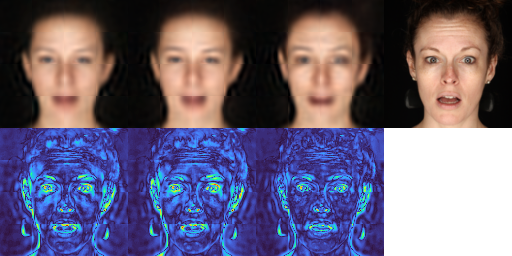}
\includegraphics[width=0.4\textwidth]{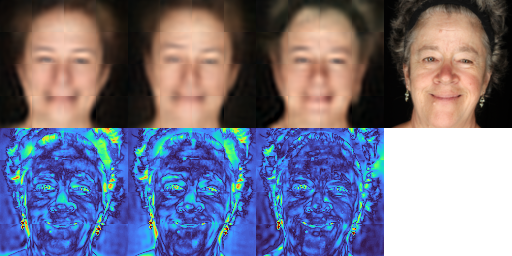}
\includegraphics[width=0.4\textwidth]{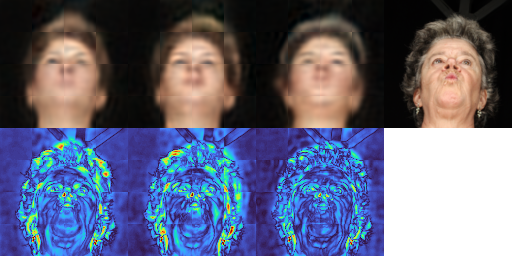}
\caption{Additional reconstruction results for test images.}
\label{fig:faces}
\end{figure*}

\begin{figure*}[hb]
    \centering
    \includegraphics[width=0.9\textwidth]{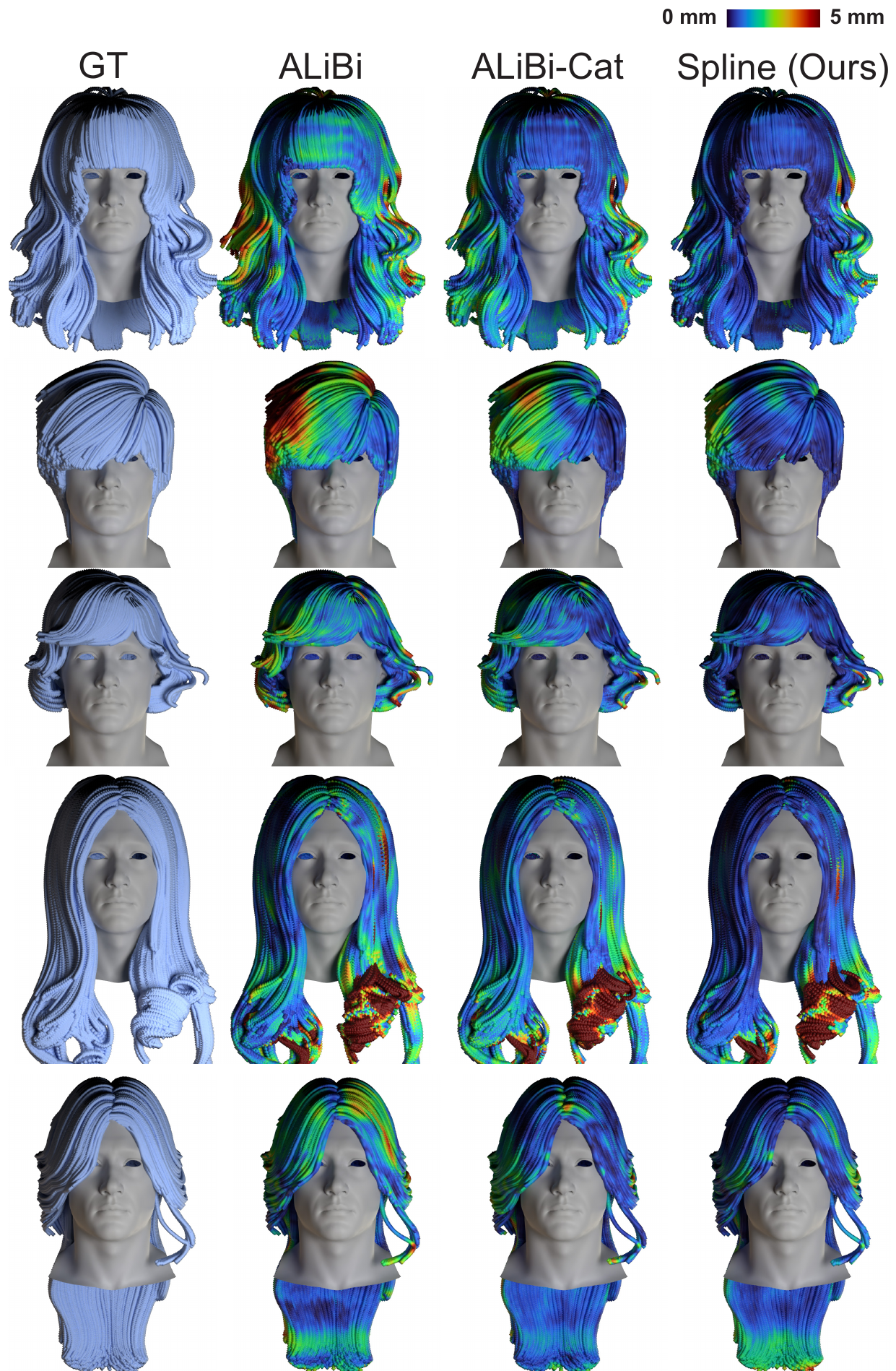}
    \caption{Additional reconstruction results for test hairstyles.}
    \label{fig:strands}
\end{figure*}

Additional results for reconstructing synthetic curves, facial images, and hair strands are shown in Figs.~\ref{fig:curves}, \ref{fig:faces}, and \ref{fig:strands}, respectively.

\end{document}